%% file: template.tex
\documentclass[research]{fcs}
\usepackage{layout}
\usepackage{subfig}
\usepackage{hyperref}
\usepackage{algorithm}
\usepackage{algorithmic}
\usepackage{subcaption}
\usepackage{subfig}
\usepackage{multirow}

\title{Reasoning Pattern Matters: \\Learning to Reason without Human Rationales}

\author[1]{Chaoxu~PANG}
\author[1,+]{Yixuan~CAO}
\author[1,+]{Ping~LUO}
\address[1]{Key Laboratory of Intelligent Information Processing, Institute of Computing Technology, Chinese Academy of Sciences.}

\corremail{\{caoyixuan, luop\}@ict.ac.cn}
\fcssetup{
  received = {month dd, yyyy},
  accepted = {month dd, yyyy},
}

\begin{abstract}
Large Language Models (LLMs) have demonstrated remarkable reasoning capabilities under the widely adopted SFT+RLVR paradigm, which first performs Supervised Fine-Tuning (SFT) on human-annotated reasoning trajectories (rationales) to establish initial reasoning behaviors, then applies Reinforcement Learning with Verifiable Rewards (RLVR) to further optimize the model using verifiable reward signals without golden rationales. However, annotating high-quality rationales for the SFT stage remains prohibitively expensive. This paper investigates when and how the cost of rationale annotations can be substantially reduced without compromising reasoning performance.  
We identify a broad class of problems, termed \emph{patterned reasoning tasks}, where the reasoning process follows a fixed, procedural solution strategy that remains consistent across all instances of the same task. Although individual instances vary in their content—such as domain knowledge, factual information, or numeric values—the solution is obtained by applying a shared reasoning pattern to instance-specific inputs. We argue that the success of SFT+RLVR on such tasks primarily stems from its ability to enable the model to internalize the underlying reasoning patterns.  
Using numerical semantic matching as a representative task, we provide two complementary lines of evidence: (i) from a \emph{cause-side} perspective, controlled experiments demonstrate that reasoning patterns—rather than the quantity or quality of rationales—are the dominant factor driving performance; and (ii) from an \emph{effect-side} perspective, forking-token analysis reveals that models trained with SFT+RLVR exhibit more task-relevant reasoning behaviors, indicating stronger alignment with the task's inherent reasoning pattern.  
Building on these insights, we propose \textbf{PARO} (\textbf{P}attern-\textbf{A}ware LLMs as \textbf{R}ationale Ann\textbf{O}tators), a simple yet effective framework that enables LLMs to generate rationales aligned with task-specific reasoning patterns \emph{without requiring human rationale annotations}. Experiments on two patterned reasoning tasks show that PARO-generated rationales achieve comparable SFT+RLVR performance to human rationales that are 10× larger. These results suggest a paradigm shift: for patterned reasoning tasks, large-scale human rationale annotations can be replaced with LLM-based automatic annotations, requiring only limited human supervision over reasoning patterns.
\end{abstract}


\begin{document}

\input{introduction}

\input{preliminaries}

\input{patterned_reason}

\input{experiment}
\input{analysis}

\input{application}
\input{related_work}

\input{conclusion}

\input{appendix}

\newpage
\bibliographystyle{fcs}
\bibliography{ref}

\begin{biography}{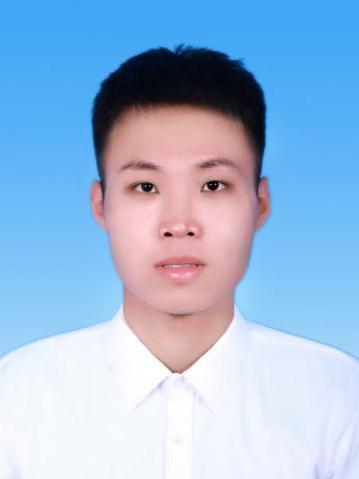}
is currently a fourth-year PhD student at the Key Laboratory of Intelligent Information Processing, Institute of Computing Technology, Chinese Academy of Sciences (ICT, CAS), under the supervision of Professor Ping Luo. Before joining ICT, he received his bachelor's degree in information and communication engineering from Beijing University of Posts and Telecommunications. His research interests lie in natural language processing, large language models, and table understanding. He has published innovative works in top-tier conferences such as ACL and EMNLP.
\end{biography}

\begin{biography}{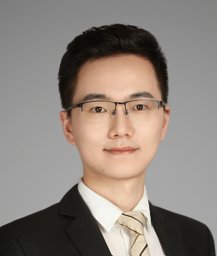}
is currently an associate professor in the Key Laboratory of Intelligent Information Processing, Institute of Computing Technology, Chinese Academy of Sciences (ICT, CAS). Before joining ICT, he received his Ph.D. in computer science from the University of Chinese Academy of Sciences (Institute of Computing Technology) in 2020. His research interests lie in natural language processing, document intelligence, and trustworthy AI. He has published papers in top-tier journals and conferences, such as KDD, WWW, CIKM, NeurIPS, and AAAI.
\end{biography}

\begin{biography}{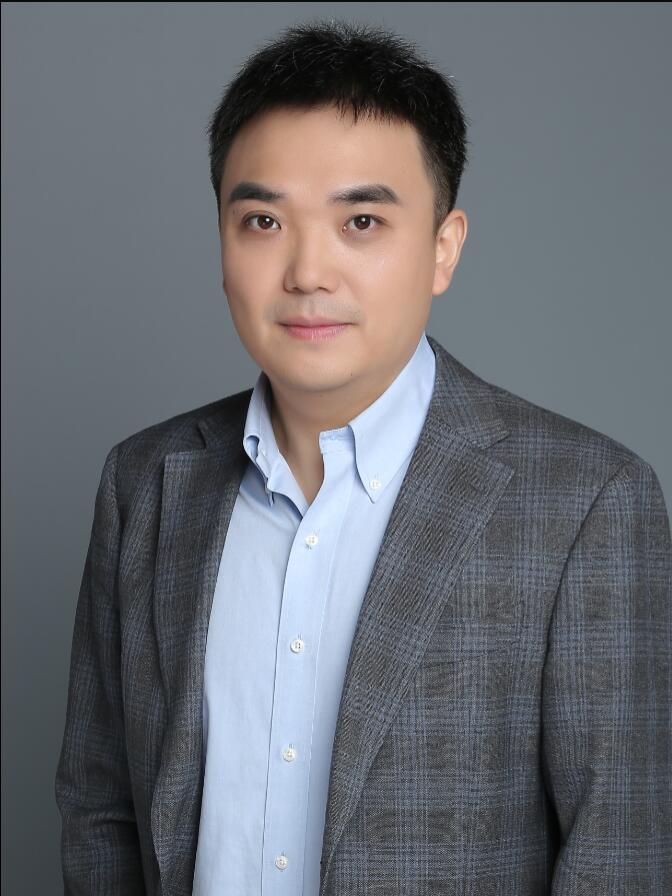}
is currently an associate professor at the Key Laboratory of Intelligent Information Processing, Institute of Computing Technology, Chinese Academy of Sciences (ICT, CAS), and the University of Chinese Academy of Sciences. Before joining ICT, he served as a senior research scientist and research manager at Hewlett-Packard Labs, China. His research interests include data mining and machine learning, with a particular focus on document AI. Dr. Luo has published over 100 research papers in top-tier journals and conferences, such as IEEE TKDE, KDD, CIKM, WSDM, ICDM, and NeurIPS. He has received several prestigious awards, including the ACM CIKM Best Student Paper Award (2012), ACM CIKM Best Paper Candidate Award (2010), SDM Best Paper Candidate Award (2010), and the Doctoral Dissertation Award from the China Computer Federation (2009).
\end{biography}

\end{document}

%% file: introduction.tex
\section{Introduction}

Large Language Models (LLMs) have recently achieved remarkable progress on complex reasoning-intensive tasks such as mathematics~\cite{shao2024deepseekmath,guo2025deepseek,chen2025bridging,yu2025dapo} and coding~\cite{guo2025deepseek,jiang2024survey}. 
A key driver of this progress is the standard two-stage training paradigm of \textbf{SFT+RLVR}. 
In the first stage, Supervised Fine-Tuning (SFT), models are exposed to high-quality reasoning trajectories (rationales) to encourage explicit reasoning behaviors. 
In the second stage, Reinforcement Learning with Verifiable Rewards (RLVR)~\cite{Lambert2024Tulu3RLVR}, reinforcement learning algorithms such as GRPO~\cite{shao2024deepseekmath} or PPO~\cite{schulman2017proximal} further optimize the model using rule-based reward signals derived from verifiable (question, answer) pairs, without requiring golden rationales. 
However, annotating large-scale, high-quality rationales for the SFT stage remains prohibitively expensive~\cite{Lambert2024Tulu3RLVR,position2025expensive}, raising a fundamental question: \emph{when and how can rationale annotation costs be reduced without compromising reasoning performance?}

We address this question for a broad class of problems that we term \textbf{patterned reasoning tasks}—tasks where the reasoning process follows a fixed, procedural solution strategy that remains consistent across all instances of the same task, while individual instances vary in their content such as domain knowledge, factual information, or numeric values. The solution is obtained by applying a shared reasoning pattern to instance-specific inputs.
Many criterion-driven problems fall into this category. Examples include text classification (e.g., topic classification~\cite{zhang2024sentiment} with detailed category definitions), verification tasks that typically follow fixed verification routines~\cite{vykopal2024fact}, and information extraction problems governed by predefined schemas~\cite{pang2023guideline,dagdelen2024structured}. Such tasks are particularly prevalent in professional domains where decision-making workflows are well-defined, such as medical diagnosis pipelines~\cite{zuo2025kg4diagnosis}, financial auditing processes~\cite{hillebrand2023improving}, and scientific information extraction~\cite{john2025human}. In each case, different inputs invoke the same decision or extraction procedure, so model performance is determined largely by how well the model learns and applies the shared pattern to variable content. 
We show more details and example tasks in Section~\ref{sec:pr_task}. 
By contrast, open-ended or adaptive reasoning tasks, such as heterogeneous mathematical problem solving~\cite{shao2024deepseekmath}, competitive programming challenges~\cite{guo2025deepseek,jiang2024survey}, and complex  planning problems~\cite{hao2025realworldplanning}, require selecting or adapting solution strategies on a per-instance basis, and therefore cannot be captured by a single fixed reasoning pattern.

We posit that the effectiveness of SFT+RLVR on patterned reasoning tasks arises from its ability to enable the model to internalize the underlying reasoning patterns. Using numerical semantic matching as a representative task, we present two complementary lines of evidence:

\begin{enumerate}
  \item \textbf{Reasoning pattern, rather than rationale quantity or quality, is the dominant factor driving SFT+RLVR performance.} 
  From a \emph{cause-side} perspective, controlled experiments reveal that: 
  (i) reducing human-annotated rationales for SFT by an order of magnitude (10× fewer samples) while preserving the reasoning pattern leads to negligible performance degradation; and 
  (ii) randomly corrupting a substantial portion of rationales (e.g., 25\%) while maintaining the pattern yields minimal impact. 
  These findings suggest that LLMs primarily learn \emph{how} to reason—the procedural pattern of thought—rather than memorizing instance-specific rationale content.

  \item \textbf{Models trained with SFT+RLVR exhibit more task-relevant reasoning behaviors, indicating stronger alignment with the task’s inherent reasoning pattern.} 
From an \emph{effect-side} perspective, we analyze model reasoning behavior through \emph{forking tokens}~\cite{wang2025beyond}—key tokens that mark decision points in reasoning trajectories and serve as indicators of reasoning pattern comprehension~\cite{wang2025beyond,cheng2025reasoning}. 
Models trained with SFT+RLVR produce forking tokens that are substantially more task-relevant, demonstrating deeper alignment with the task’s inherent reasoning pattern. 
In contrast, models trained solely with RLVR or hint-based methods~\cite{liu2025uft} tend to generate generic discourse connectors (e.g., ``but,'' ``because''), reflecting a lack of focus on the core reasoning patterns essential for task completion.

\end{enumerate}

Building on these insights, we introduce \textbf{P}attern-\textbf{A}ware LLMs as \textbf{R}ationale Ann\textbf{O}tators (\textbf{PARO}), a simple yet effective framework that explicitly instructs strong LLMs to generate rationales following task-specific reasoning patterns. 
We evaluate PARO on two representative patterned reasoning tasks—numerical semantic matching and transaction purpose classification. 
PARO achieves comparable performance to human rationale datasets that are 10× larger, while not requiring human rationale annotations, and outperforms approaches that distill internal reasoning trajectories from large reasoning models such as Qwen3-235B-A22B-Thinking~\cite{qwen3technicalreport}.

Taken together, our findings reveal that for patterned reasoning tasks, the central challenge lies not in collecting more high-quality rationales, but in defining and enforcing clear reasoning patterns. 
This insight preserves model performance while dramatically reducing annotation costs, providing a practical and scalable pathway for reasoning supervision in LLMs.

%% file: preliminaries.tex
\section{Preliminaries}

Reinforcement Learning from Verifiable Rewards (RLVR)~\cite{Lambert2024Tulu3RLVR} has emerged as a widely adopted paradigm for enhancing the reasoning capabilities of LLMs. Prior to RLVR, models are typically warm-started with a \textbf{Supervised Fine-Tuning (SFT)} stage, which encourages explicit, human-readable reasoning trajectories. 

Concretely, the SFT stage requires a small collection of question--rationale--answer triples $\mathcal{D}_r = \{(q, r, a)\}$, where $q$, $r$, and $a$ denote the question, rationale, and answer, respectively. To ensure high quality, rationales are typically annotated by human experts. The SFT objective fine-tunes the model by maximizing the likelihood of generating a concatenated rationale--answer sequence $f_e(r,a)$ conditioned on the question $q$:
\begin{equation}
    \theta^{(1)} = \arg\max_{\theta} \mathbb{E}_{(q,r,a) \sim \mathcal{D}_r} 
    \left[ \log \pi_\theta\!\left(f_e(r,a) \mid q\right) \right].
\end{equation}

Subsequently, the \textbf{RLVR stage} initializes from $\theta^{(1)}$ and further optimizes the model on a larger dataset of question--answer pairs $\mathcal{D}_d = \{(q,a)\}$ using the RLVR objective:
\begin{equation}
    \max_{\theta} \; \mathbb{E}_{(q,a) \sim \mathcal{D}_d}\;
    \mathbb{E}_{y \sim \pi_\theta(\cdot \mid q)} \big[ R_{\text{RLVR}}(q,a,y) \big],
\end{equation}
where $y$ denotes the response sampled from the policy $\pi_\theta$.
The core principle underlying the RLVR objective is straightforward: the model receives reward only when its generated response is \emph{verifiably correct}. Formally, the RLVR reward function can be expressed as:
\begin{equation}
    R_{\text{RLVR}}(q,a,y) = 
    v(y,a) - \beta \, \text{KL}\!\big(\pi_\theta(y \mid q)\,\|\,\pi_{\text{ref}}(y \mid q)\big),
\end{equation}
where $v$ is a verifiable reward function. A commonly used instantiation compares the extracted final answer from $y$ against the ground truth $a$:
\begin{equation}
    v(y,a) =
    \begin{cases}
        1, & \text{if } \text{extract}(y) = a, \\
        0, & \text{otherwise}.
    \end{cases}
\end{equation}
Here, $\text{extract}(\cdot)$ denotes an answer-extraction function that parses the final prediction from the generated rationale. This objective is typically optimized using policy-gradient algorithms such as PPO~\cite{schulman2017proximal} or GRPO~\cite{shao2024deepseekmath}.

%% file: patterned_reason.tex
\section{Patterned Reasoning Tasks}\label{sec:pr_task}

In this section, we formally characterize patterned reasoning tasks and distinguish them from adaptive reasoning tasks, establishing the conceptual framework that defines the scope of our study.

\subsection{Definition and Characteristics}

We define \textbf{patterned reasoning tasks} as tasks where the overall reasoning process follows a fixed, procedural solution strategy across all instances. While the reasoning content may vary between instances and require diverse knowledge (e.g., commonsense, domain-specific expertise, or deductive logic), the underlying \emph{reasoning pattern} remains invariant. 
In contrast, \textbf{adaptive reasoning tasks} require the reasoning process to flexibly adapt to instance-specific characteristics, resulting in diverse solution strategies that cannot be captured by a single fixed pattern.

Formally, a patterned reasoning task can be characterized by:
\begin{itemize}
    \item \textbf{Stable reasoning pattern} \(\mathcal{P}\): a consistent procedural framework applicable across all task instances.
    \item \textbf{Instance-specific content} \(\mathcal{C}_i\): variable knowledge, facts, or values required for each instance \(i\).
    \item \textbf{Pattern execution function} \(f(\mathcal{P}, \mathcal{C}_i) \rightarrow y_i\): the application of \(\mathcal{P}\) to \(\mathcal{C}_i\) to produce output \(y_i\).
\end{itemize}

The key distinction lies in whether the reasoning pattern \(\mathcal{P}\) depends on the instance. For patterned reasoning tasks, \(\mathcal{P}\) remains consistent across instances, whereas adaptive reasoning tasks require \(\mathcal{P}_i\) to be flexibly chosen or adapted for each instance \(i\).

\subsection{Task Categories}

We provide representative examples of the two reasoning tasks to illustrate their typical task types.

\paragraph{Adaptive reasoning tasks.}
\begin{itemize}
    \item \textbf{Mathematical problem solving}~\cite{shao2024deepseekmath}: Diverse problems requiring distinct solution strategies.
    \item \textbf{Programming tasks}~\cite{guo2025deepseek,jiang2024survey}: Coding problems demanding varied algorithmic designs and implementation approaches.
    \item \textbf{Planning tasks}~\cite{hao2025realworldplanning}: Sequential decision-making problems where different initial states or goals necessitate distinct strategic approaches.
\end{itemize}

\paragraph{Patterned reasoning tasks.}
\begin{itemize}
    \item \textbf{Criterion-driven problems}: Tasks with predefined categorical definitions or explicit discrimination criteria:
    \begin{itemize}
        \item \emph{Classification problems}: Tasks such as sentiment analysis~\cite{zhang2024sentimentLLM}, topic classification~\cite{gretz2023zeroshot}, and intent recognition~\cite{arora2024intentLLMs}, governed by fixed label sets and decision criteria. These tasks rely on consistent feature patterns and boundary conditions to assign inputs to predefined categories.
        \item \emph{Verification problems}: Tasks such as fact-checking~\cite{li2024self,leippold2025automated}, following clear verification protocols. The criteria involve cross-referencing claims against reliable sources and applying logical consistency checks to determine truthfulness.
        \item \emph{Extraction problems}: Tasks such as relation extraction~\cite{wan2023gpt}, event extraction~\cite{pang2023guideline}, and table information extraction~\cite{pang2024uncovering,zhao2023investigating}, which follow consistent annotation schemas. These tasks operate under predefined entity types, relationship categories, and structural templates that guide the identification and extraction process.
    \end{itemize}
    \item \textbf{Deductive reasoning}: Tasks applying established rules to derive conclusions:
    \begin{itemize}
        \item \emph{Logical reasoning}: Tasks such as symbolic reasoning~\cite{pan2023logic}.
        \item \emph{Mathematical calculations}: Tasks such as geometric or algebraic computations~\cite{weng2025geosketch}.
        \item \emph{Algorithmic simulation}: Tasks such as executing well-defined stepwise procedures~\cite{ma2024llm}.
    \end{itemize}
\end{itemize}

Patterned reasoning tasks are particularly prevalent in professional and specialized domains, where decision-making workflows are typically well-defined and domain experts follow relatively standardized procedures (e.g., medical diagnosis pipelines~\cite{zuo2025kg4diagnosis}, financial auditing processes~\cite{hillebrand2023improving}, or scientific information extraction~\cite{john2025human}). 

In this work, we focus on two representative patterned reasoning tasks from the financial domain: numerical semantic matching and transaction purpose classification. We construct new datasets with carefully annotated labels and rationales. Unlike many widely studied benchmarks, these tasks have limited exposure in current pre-training corpora, which helps mitigate both data contamination (where models memorize benchmark samples) and task contamination (where models implicitly learn task-specific strategies from repeated exposure to public benchmarks). We present detailed analyses of the task definitions and their underlying reasoning patterns in the following subsections.

\subsection{Numerical Semantic Matching}\label{sec:nsm}

\begin{figure*}[!tb]
    \centering
    \includegraphics[width=0.85\linewidth]{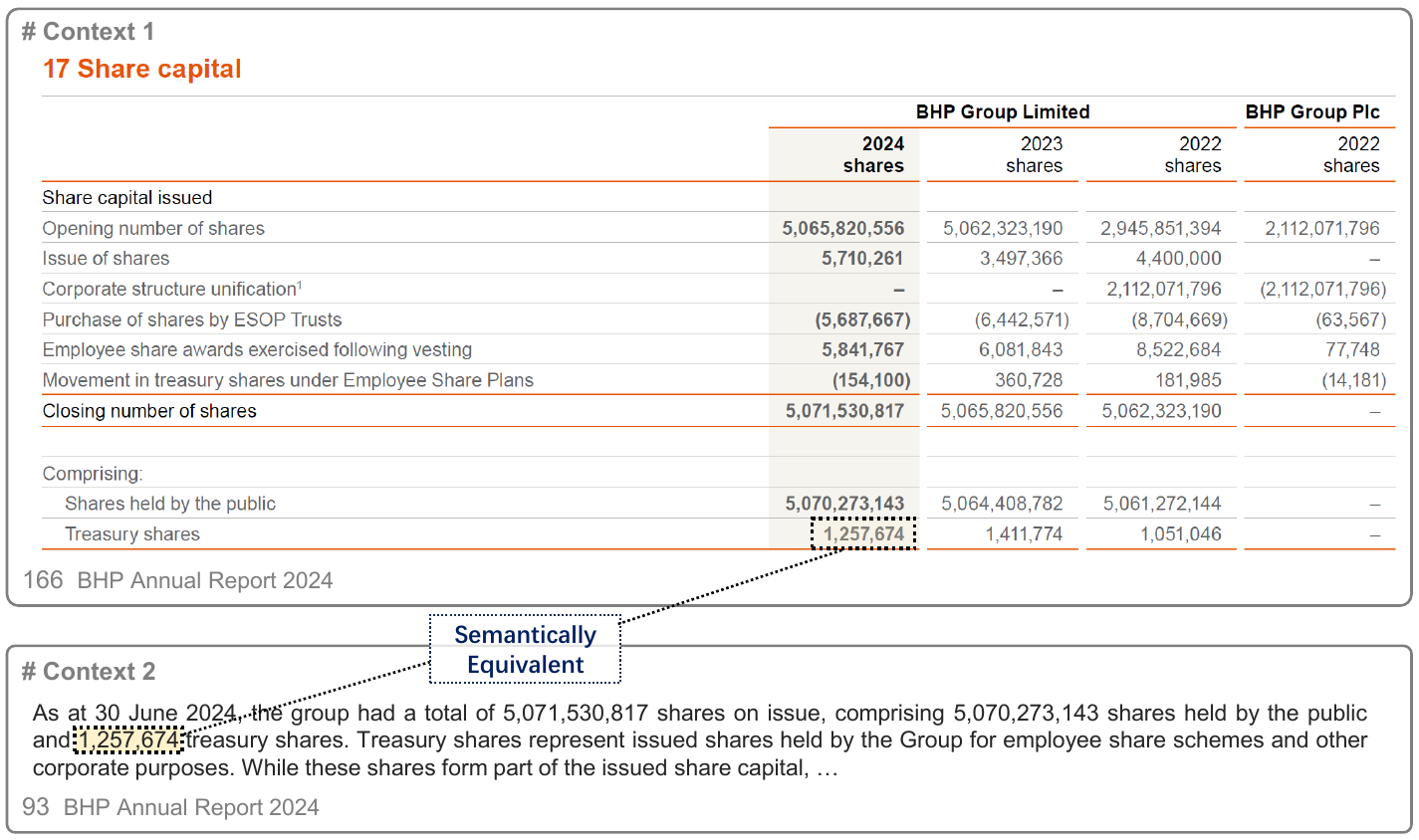}
    \caption{Two document excerpts from the BHP Annual Report 2024. The numerical mentions highlighted in dashed boxes are semantically equivalent.}
    \label{fig:intro_example}
\end{figure*}

Numerical mentions are pervasive in documents such as financial reports, regulatory filings, and web pages, where they play a central role in statistical reporting and disclosure. \textbf{Numerical Semantic Matching (NSM)} is the task of determining whether two numerical mentions are \emph{semantically equivalent}, i.e., whether they refer to the same underlying numerical fact. 

As shown in Figure~\ref{fig:intro_example}, two distinct mentions may appear in different parts of a company's annual report, yet both correspond to the identical fact that \textit{the treasury share component within BHP Group Limited's closing number of shares at the end of fiscal year 2024 is 1,257,674}. Accurately detecting such equivalence has broad applications such as fact checking~\cite{pang2025document}, knowledge base construction~\cite{zheng2023survey}, and numerical reasoning~\cite{ran2019numnet,akhtar2023exploring,wang2024encore}.

\paragraph{Task formulation.} Formally, given two numerical mentions $n_1$ and $n_2$ with their associated contexts $c_1$ and $c_2$, the goal of NSM is to determine whether $n_1$ and $n_2$ refer to the same fact (binary classification). Each context may be a paragraph or a table, and we standardize them into textual form (with tables linearized using Markdown).

\paragraph{Reasoning pattern.} Solving NSM can be generally decomposed into four steps:
\begin{enumerate}
    \item \textbf{Numerical grounding:} Locate each numerical mention in the given context.  
    \item \textbf{Semantic interpretation:} Identify the semantics of the numerical mention within its context (e.g., time, subject, indicator).
    \item \textbf{Entity alignment:} Link contextual references to consistent entities or events (e.g., ``BHP Group Limited'' vs.\ ``the company''; ``FY2024'' vs.\ ``as of June 2024'').  
    \item \textbf{Equivalence decision:} Compare the semantic frames of both mentions and decide whether they denote the same fact.  
\end{enumerate}
This reasoning process demonstrates that NSM goes beyond simple string matching or numerical comparison, requiring the integration of quantitative values with domain-specific contextual cues to determine semantic equivalence.

\subsection{Transaction Purpose Classification}\label{sec:tpc}

Banking systems generate vast volumes of transaction data, and accurately identifying the underlying purpose of each transaction is crucial for compliance auditing, financial analysis, and fraud detection. The \textbf{Transaction Purpose Classification (TPC)} task addresses this need by assigning a single transaction record to one of 62 predefined categories, covering both corporate-related and personal-related purposes. 

\paragraph{Task formulation.}  
Formally, given a structured record consisting of:  
(i) account holder type (enterprise/individual),  
(ii) transaction direction (credit/debit),  
(iii) free-text transaction memo,  
(iv) counterparty identity, and  
(v) contextual metadata (e.g., time, channel, amount),  
the task is to predict the correct purpose category from a fixed taxonomy of 62 classes.

\paragraph{Reasoning pattern.} Solving TPC can be generally decomposed into four steps:
\begin{enumerate}
    \item \textbf{Identify key attributes:} Determine account holder type and transaction direction.  
    \item \textbf{Extract salient cues:} Parse informative keywords, amounts, organizations, and dates from the memo and counterparty fields.  
    \item \textbf{Apply taxonomy rules:} Match the extracted cues to the taxonomy, applying priority rules for authoritative signals (e.g., payments to tax authorities, legal institutions, or banks).  
    \item \textbf{Finalize decision:} Select the most plausible category and, if required, generate a concise rationale citing the decisive cues.  
\end{enumerate}

More details of the TPC dataset and taxonomy are provided in Appendix~D.

%% file: experiment.tex
\section{Pilot Comparison Experiments}
In this section, we use the NSM task as a case study to investigate the following research question: \textit{Does SFT+RLVR achieve the best performance in incentivizing the reasoning capabilities of LLMs?}

\subsection{Experimental Setups}
\subsubsection{Dataset}\label{sec:nsm_data}
\noindent \textbf{Training set}. We collect 110k numerical semantic matching samples from 544 annual reports of Chinese companies. Annual reports provide comprehensive overviews of companies' yearly performance, operations, and future outlook, serving to inform shareholders and stakeholders. Our dataset encompasses substantial diversity across two key dimensions: (1) \textit{Industry coverage}: The companies span multiple sectors including finance, manufacturing, technology, and real estate; (2) \textit{Temporal coverage}: The reporting periods range from 2018 to 2024, capturing temporal diversity across varying market conditions and economic cycles. 
For each sample, we annotate the ground-truth answer following the protocol established in previous work~\cite{li2020cracking}. Additionally, we engage 8 professional annotators to provide \textbf{rationales} for 10k samples, with subsequent validation performed by 2 financial experts. Statistical analysis reveals that the first-round annotation accuracy reaches approximately 97\%, which is further improved after expert validation. We denote these 10k samples with rationales as \textbf{RatQA-10k}, while the remaining 100k samples without rationales are denoted as \textbf{DirQA-100k}.

\noindent \textbf{Test set}. Following a similar data construction pipeline, we collect 20k samples each from annual reports and IPO prospectuses to form our evaluation datasets. Unlike annual reports, IPO prospectuses contain comprehensive information about a company's financials, business model, and risk factors to ensure transparency and regulatory compliance during public offerings. The IPO prospectus evaluation set serves as a more challenging cross-domain benchmark, better reflecting the model's generalization capabilities across different document types and contexts.

More details about the dataset construction and statistics are provided in Appendix~A.

\subsubsection{Metrics}
We adopt accuracy, precision, recall, and F1-score as evaluation metrics, following prior work~\cite{li2020cracking,pang2025document}. Specifically, given a set of golden semantically equivalent numerical pairs \(g\) and a set of predicted pairs \(p\), we define the metrics as follows:
{\setlength{\jot}{8pt}
\begin{align}
\text{Precision (P)} & = \frac{|g \cap p|}{|p|}, \\
\text{Recall (R)}    & = \frac{|g \cap p|}{|g|}, \\
\text{F1-score}      & = \frac{2 \cdot \text{P} \cdot \text{R}}{\text{P} + \text{R}}.
\end{align}
}

\subsubsection{Implementation Details}\label{sec:implement_detail}
We select Qwen3-8B~\cite{yang2025qwen3} as our backbone model due to its exceptional Chinese language understanding capabilities~\cite{galileo2025qwen_chinese_advantage} and its popularity as a foundation for RLVR~\cite{wang2025beyond,wu2025thought}. 

For SFT, we employ the Hugging Face Transformers library~\cite{wolf2020transformers} and DeepSpeed ZeRO~\cite{rajbhandari2020zero} for efficient distributed training. The model is trained for 2 epochs with a learning rate of 2e-5, a batch size of 20 per GPU. We use a cosine annealing scheduler with warmup for the first 1\% of training steps. The maximum gradient norm is clipped at 1.0 to ensure training stability.
For RLVR, we employ VERL~\cite{sheng2024hybridflow} for efficient distributed training. We set the temperature to 1.0 for diverse rollout sampling and sample 16 responses for each prompt during the sampling phase. We use a constant learning rate of 1e-6 with a training batch size of 192. 

All training procedures are conducted on a cluster of 24 H100 GPUs with 80GB memory each. We utilize vLLM~\cite{kwon2023efficient} for efficient inference during evaluation. For both training and inference, the maximum input and output sequence length is set to 4096 and 1024 tokens, respectively. 

\subsubsection{Baselines}
We evaluate SFT+RLVR against the following training strategies:

\begin{itemize}

    \item \textbf{SFT-direct}: Supervised fine-tuning directly on DirQA-100k, i.e., training only on answers without rationales. This is the standard SFT baseline, which reduces rationale annotation cost but does not explicitly guide the reasoning process.

    \item \textbf{SFT-rationales}: Supervised fine-tuning only on \textsc{RatQA-10k}, where the model is trained to generate rationales before predicting the final answers. This provides explicit reasoning supervision but requires costly rationale annotations.

    \item \textbf{pure-RLVR}~\cite{guo2025deepseek}: Reinforcement learning with verifiable rewards, optimized purely on \textbf{RatQA-10k}. The model is trained only with correctness feedback on final answers, without any rationale supervision.

    \item \textbf{UFT}~\cite{liu2025uft}: unified fine-tuning, which concatenates rationales as hints into the prompts and applies curriculum learning to dynamically control the proportion and length of hints. This guides the model toward correct reasoning paths while maintaining flexibility.
\end{itemize}

\subsection{Experimental Results}

\begin{table*}[ht]
\centering
\fontsize{8.6}{10.0}\selectfont
\renewcommand{\arraystretch}{1.2}
\setlength{\tabcolsep}{5pt}
\begin{tabular}{lcccccccccccc}
\toprule 
\multirow{2}{*}{\textbf{Strategy}} & \multirow{2}{*}{\textbf{RatQA-10k}} & \multirow{2}{*}{\textbf{DirQA-100k}} & \multicolumn{4}{c}{\textbf{IPO Prospectus}} & \multicolumn{4}{c}{\textbf{Annual Report}} & \multicolumn{2}{c}{\textbf{Average}} \\
\cmidrule(lr){4-7} \cmidrule(lr){8-11} \cmidrule(lr){12-13}
& & & \textbf{Acc.} & \textbf{P.} & \textbf{R.} & \textbf{F1.} & \textbf{Acc.} & \textbf{P.} & \textbf{R.} & \textbf{F1.} & \textbf{Acc.} & \textbf{F1.} \\
\midrule
SFT-direct     &            & \checkmark & 76.8 & 51.7 & 33.8 & 40.8 & 82.8 & 64.2 & 63.0 & 63.6 & 79.8 & 52.2 \\
SFT-rationales & \checkmark &  & 78.0 & 53.5 & 58.2 & 55.7 & 80.3 & 58.3 & 60.4 & 59.4 & 79.2 & 57.6 \\
pure-RLVR~\cite{guo2025deepseek}   &          & \checkmark & 87.1 & 75.2 & 68.2 & 71.5 & 89.1 & 77.0 & 77.1 & 77.1 & 88.1 & 74.3 \\
UFT~\cite{liu2025uft} & \checkmark & \checkmark & 87.5 & 76.6 & 68.3 & 72.2 & 90.7 & 83.5 & 75.9 & 79.5 & 89.1 & 75.9 \\
SFT+RLVR      & \checkmark & \checkmark & \textbf{88.3} & \textbf{79.2} & \textbf{68.8} & \textbf{73.6} & \textbf{92.3} & \textbf{87.1} & \textbf{79.6} & \textbf{83.2} & \textbf{90.3} & \textbf{78.4} \\
\cmidrule(lr){1-13}
\textit{Controlled Experiments} & & & & & & & & & & & & \\
SFT+RLVR & 1k & \checkmark & 87.7 & 77.7 & 67.3 & 72.1 & 91.8 & 85.2 & 79.5 & 82.3 & 89.8 & 77.2 \\
SFT+RLVR & 1k, 25\% wrong & \checkmark & 88.2 & 80.6 & 66.3 & 72.7 & 91.7 & 82.7 & 82.4 & 82.6 & 90.0 & 77.7 \\
\bottomrule
\end{tabular}
\caption{Comparison of different training strategies.}
\label{tab:strategy-comparison}
\end{table*}

Table~\ref{tab:strategy-comparison} summarizes the results across different training strategies on the numerical semantic matching task. Several key observations emerge:

\textbf{Rationales provide substantial benefits over direct supervision.}  
The SFT-rationales baseline, trained on only 10k samples, achieves 79.2\% average accuracy and 57.6\% F1, whereas SFT-direct, trained on 100k samples, only reaches 79.8\% accuracy and 52.2\% F1. Despite using ten times less data, SFT-rationales delivers comparable accuracy and notably higher F1, demonstrating the effectiveness of explicit reasoning supervision. This highlights that NSM requires reasoning rather than surface-level pattern learning, and well-annotated rationales significantly improve model generalization.

\textbf{SFT+RLVR achieves the strongest overall performance.}  
Among reinforcement-based strategies, pure-RLVR surpasses both SFT baselines by leveraging verifiable reward signals. UFT further improves performance by leveraging rationales as hints and applying curriculum learning strategies. However, SFT+RLVR achieves the best results overall, with 90.3\% average accuracy and 78.4\% F1, consistently outperforming both pure reinforcement and unified hint-based training.  

\textbf{Cross-domain consistency.}  
The evaluation across both IPO Prospectus and Annual Report domains reveals remarkably consistent relative performance rankings, with SFT+RLVR maintaining its superiority in both domains (IPO: 88.3\% accuracy, 73.6\% F1; Annual Report: 92.3\% accuracy, 83.2\% F1). This cross-domain consistency validates the robustness of SFT+RLVR and suggests that the reasoning capabilities developed through SFT+RLVR generalize effectively across different financial document types. The consistent performance gaps between methods across domains further confirm that the observed improvements are not artifacts of specific data characteristics but reflect genuine enhancements in NSM reasoning ability.

\textbf{Conclusion.} Taken together, these results highlight two key insights: (i) reasoning capabilities play a pivotal role in numerical semantic matching, and (ii) integrating rationale-based supervised fine-tuning with reinforcement learning (SFT+RLVR) provides the most effective strategy among all compared methods for stimulating and enhancing the model’s reasoning ability.

%% file: analysis.tex
\begin{figure*}[t]
    \centering
    \subfloat[Impact of rationale quantity on the performance of SFT and RLVR stages. Results are reported on the annual report subset.]
    {\includegraphics[width=0.43\textwidth]{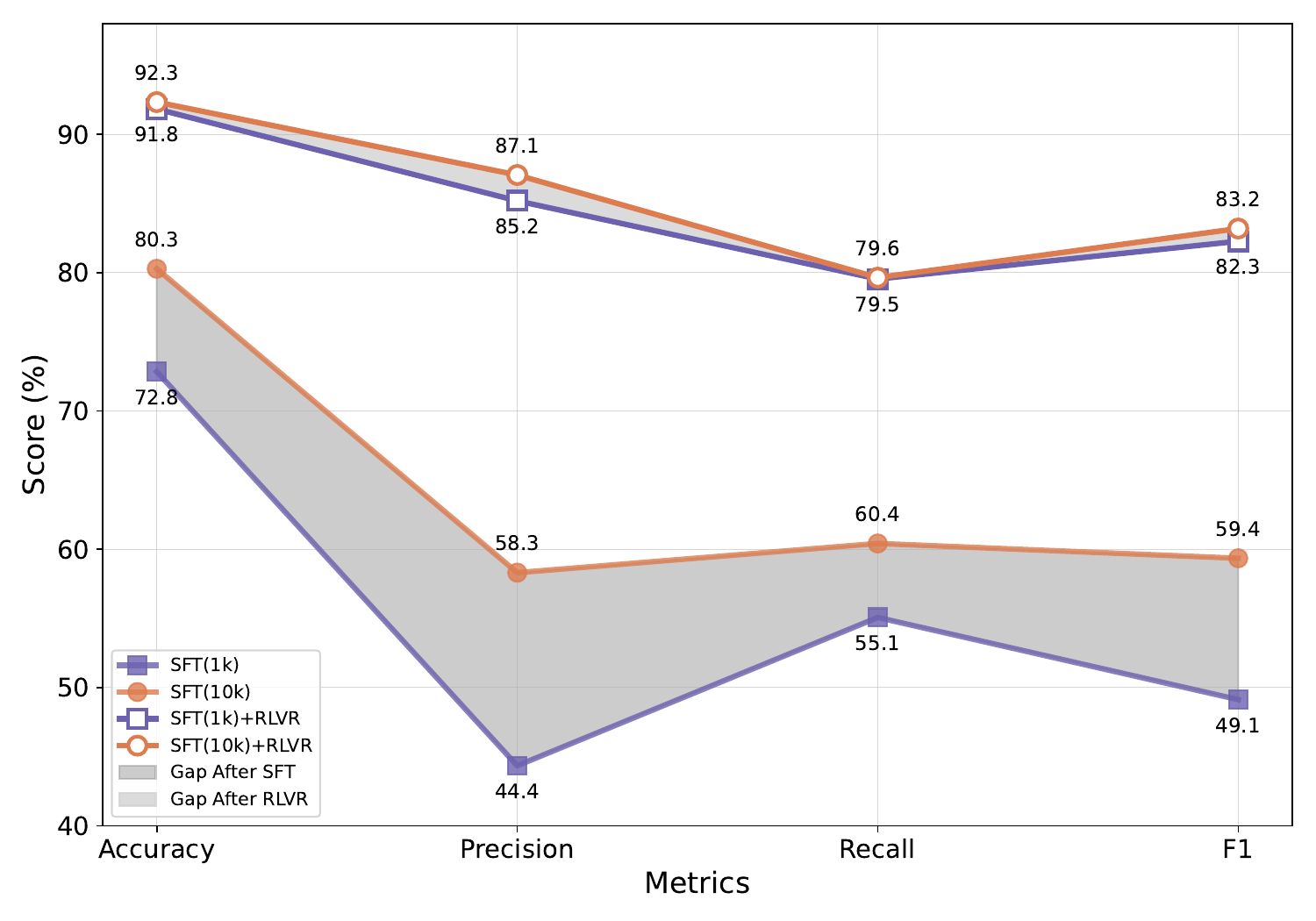}
    \label{fig:datasize-sft-rl}}
    \hspace{0.04\textwidth}
    \subfloat[Impact of rationale quality on the performance of SFT and RLVR stages. Results are reported on the annual report subset.]
    {\includegraphics[width=0.43\textwidth]{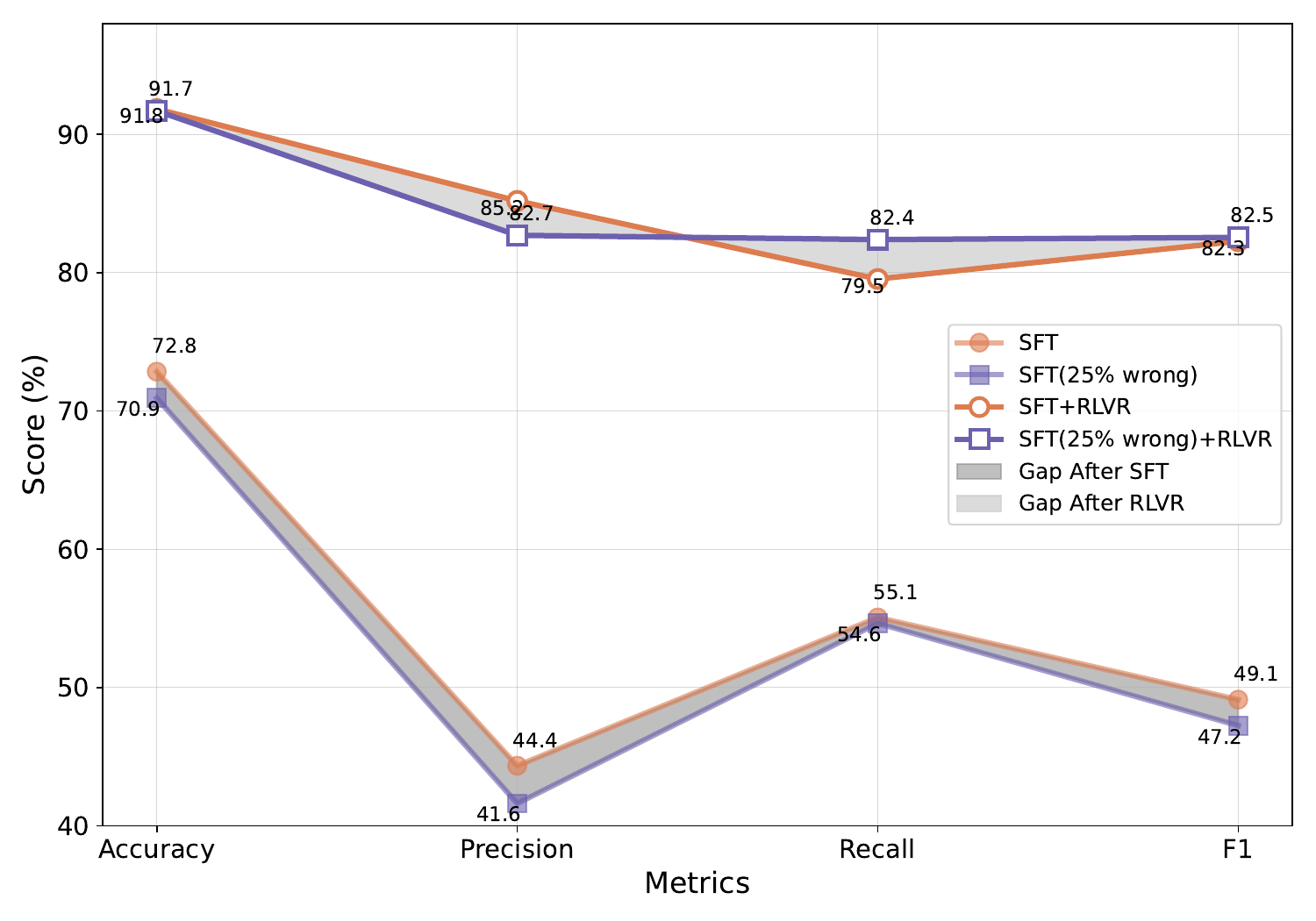}
    \label{fig:label-correct-sft-rl}}
    \caption{Performance comparison of SFT and RLVR stages under varying rationale quantity (a) and quality (b).}
    \label{fig:ablations-sft-rl}
\end{figure*}

\section{Analysis of the Importance of Reasoning Patterns}

In this section, we present a comprehensive analysis to explain why SFT+RLVR substantially outperforms other training strategies. 
We hypothesize that SFT+RLVR enables models to more effectively acquire reasoning patterns by providing direct supervision on rationales during fine-tuning. 
To validate this hypothesis, we provide two complementary lines of evidence: 
(1) from a \emph{cause-side} perspective, through controlled experiments (Section~\ref{sec:abl_study}), we demonstrate that it is the reasoning pattern—rather than the mere quantity or quality of rationales—that plays the decisive role in the success of SFT+RLVR; and 
(2) from a \emph{effect-side} perspective, through forking token analysis (Section~\ref{sec:fork_token}), we show that models trained with SFT+RLVR capture task-specific reasoning patterns more accurately, leading to more focused problem-solving behavior.

\subsection{\textbf{Evidence 1: Effect of Rationale Quantity and Quality}}\label{sec:abl_study}

In this section, we design controlled experiments to demonstrate that reasoning pattern serves as the key factor when leveraging rationales, outweighing two alternative factors: the quantity (Section~\ref{sec:task_specific_knowledge}) and quality (Section~\ref{sec:label_correctness}) of rationales.

\subsubsection{Reasoning Pattern over Rationale Quantity}\label{sec:task_specific_knowledge}

To achieve this, we reduce the size of the RatQA-10k dataset by 10-fold, randomly retaining only 1k out of 10k training samples. Previous studies~\cite{jiang2025predicting} have shown that training dataset size is highly correlated with the volume of task-specific knowledge acquired. This reduction results in significantly lower task-specific knowledge while preserving the overall reasoning pattern present in each training sample.

We employ the same training recipe as SFT+RLVR described in Section~\ref{sec:implement_detail}. The results of SFT+RLVR (only 1k rationale samples) are presented in Table~\ref{tab:strategy-comparison}. The performance remains comparable to the full SFT+RLVR, with F1 decreasing by merely 1.2 points. This is particularly notable when compared to other reasoning-enhanced methods, which lag significantly behind SFT+RLVR by 2.5-20.8 points. These results indicate that training dataset size is not the primary factor driving the performance gains of SFT+RLVR, thereby supporting our hypothesis that reasoning patterns are the crucial component.

\subsubsection{Reasoning Pattern over Rationale Quality}\label{sec:label_correctness}

Previous work~\cite{min2022rethinking} has indicated that golden demonstrations are not strictly required for in-context learning—randomly replacing labels in demonstrations barely hurts the performance across a range of classification and multiple-choice tasks. In this section, we investigate whether this phenomenon extends to SFT+RLVR: specifically, whether the correctness of rationales significantly impacts performance. 
To achieve this, we randomly replace 25\% of human-annotated rationales with incorrect rationales while maintaining the underlying reasoning pattern unchanged. We prompt GPT-4.1\footnote{gpt-4.1-2025-04-14} to modify 25\% of manually-annotated rationales into incorrect rationales. The detailed prompt is provided in Appendix C. 

We employ the same training setup as SFT+RLVR described in Section~\ref{sec:implement_detail}. The results of SFT+RLVR (1k, 25\% wrong) are shown in Table~\ref{tab:strategy-comparison}. Remarkably, its performance remains comparable to that of the full SFT+RRFT, with F1 decreasing by only 0.7 points. 
Interestingly, we observe that SFT+RLVR (1k, 25\% wrong) slightly outperforms SFT+RLVR (1k). We hypothesize that this improvement arises because model-modified responses introduce greater response diversity than human rationales, thereby enhancing the overall performance~\cite{matsutani2025rl}. We leave a detailed investigation of this phenomenon for future work.

These results demonstrate that the quality of rationales is not the primary factor driving the performance gains of SFT+RLVR. Instead, they provide compelling evidence that reasoning patterns serve as the key component, consistent with findings from the in-context learning literature, and further support our central hypothesis.

\subsubsection{Analysis of Performance Gaps Between Training Phases}

To better understand the roles of SFT and RLVR stages, we conduct a analysis of intermediate performance under different rationale data conditions from Section~\ref{sec:task_specific_knowledge} and Section~\ref{sec:label_correctness}. Figure~\ref{fig:datasize-sft-rl} and Figure~\ref{fig:label-correct-sft-rl} present the performance comparison after each training phase across different evaluation metrics.

The results reveal two key insights about our SFT+RLVR training pipeline:

\textbf{SFT stage: Quantity and quality of rationales are critical.} During the SFT stage, both the quantity and quality of rationales significantly impact model performance. The 10k SFT model substantially outperforms the 1k SFT model across all metrics (e.g., 59.4\% vs. 49.1\% F1 score), demonstrating that abundant rationale data provides richer task-relevant knowledge for establishing stronger initial reasoning capabilities. A similar pattern emerges in Figure~\ref{fig:label-correct-sft-rl} when comparing models trained with correct versus incorrect rationales (25\% wrong), where data quality directly translates to a 2\% F1 score degradation. These substantial gaps underscore the importance of rationale quantity and quality in building a solid foundation for reasoning.

\textbf{RLVR Stage: Performance Convergence via Self-Refining Reasoning.} 
After RLVR training, the performance gaps across settings almost vanish, with final models achieving highly similar results, differences of only 0.9\% and 0.2\% in F1 score for the rationale quantity and quality experiments, respectively. 
This convergence suggests that RLVR effectively compensates for initial weaknesses by acquiring task-relevant knowledge from large-scale (question, answer) pairs, even when starting from suboptimal SFT baselines. 
The underlying mechanism behind this robustness is that once the SFT stage establishes basic reasoning patterns—regardless of data limitations or quality issues—the model can continue to self-improve during RLVR by generating diverse rationales, reinforcing reasoning paths that lead to correct answers, and suppressing erroneous ones.

This complementary learning dynamic highlights the synergistic relationship between SFT and RLVR stages: \textbf{SFT provides essential reasoning patterns and templates, while RLVR enables scalable knowledge acquisition through self-exploration of the reasoning space.} This property makes the approach robust to limited availability of high-quality rationale data, a common challenge in real-world applications.

\subsection{\textbf{Evidence 2: Analysis on Reasoning Behaviors}}\label{sec:fork_token}

\begin{figure}[!tb]
    \centering
    \includegraphics[width=0.95\linewidth]{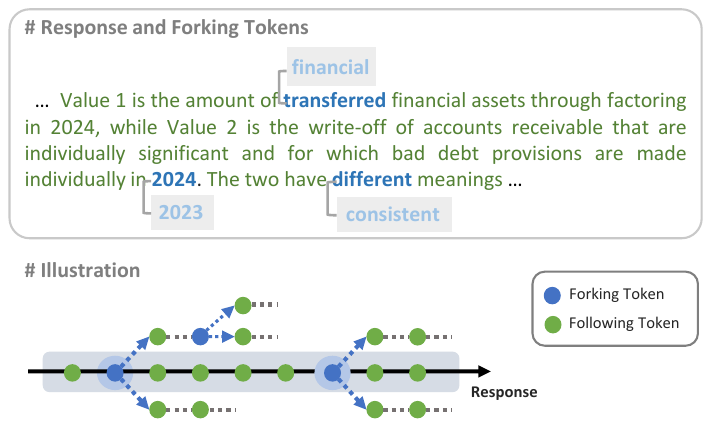}
    \caption{An Example of forking tokens in LLM-generated responses.}
\label{fig:fork_token}
\end{figure}

In the previous section, we provided evidence from an \textit{cause-side} perspective through controlled experiments. However, this analysis still lacks the \textit{effect-side} perspective, focusing on how reasoning patterns are concretely manifested in model responses. 
To address this gap, we identify and analyze the forking tokens~\cite{wang2025beyond} within model responses—critical decision points that fundamentally shape the reasoning trajectory and final outcomes.
As illustrated in Figure~\ref{fig:fork_token}, the LLM exhibits three pivotal decision points (i.e., forking tokens) when generating this response: 
(1) \textit{transferred} vs \textit{financial}, which determines the semantic interpretation of the accounting subject for value 1; 
(2) \textit{2024} vs \textit{2023}, which specifies the temporal context of value 2; and 
(3) \textit{different} vs \textit{consistent}, which governs the ultimate decision on semantic equivalence between the two values. 
In contrast, the model demonstrates high certainty at other token positions, which exert minimal influence on the final outcome. 
Forking tokens thus serve as crucial indicators of the model’s underlying reasoning patterns. 
By analyzing these forking tokens, we can obtain deeper insights into the model’s reasoning behaviors and decision-making processes.

\begin{figure*}[!tb]
    \centering
    \begin{minipage}{0.32\textwidth}
        \centering
        \includegraphics[width=0.9\linewidth]{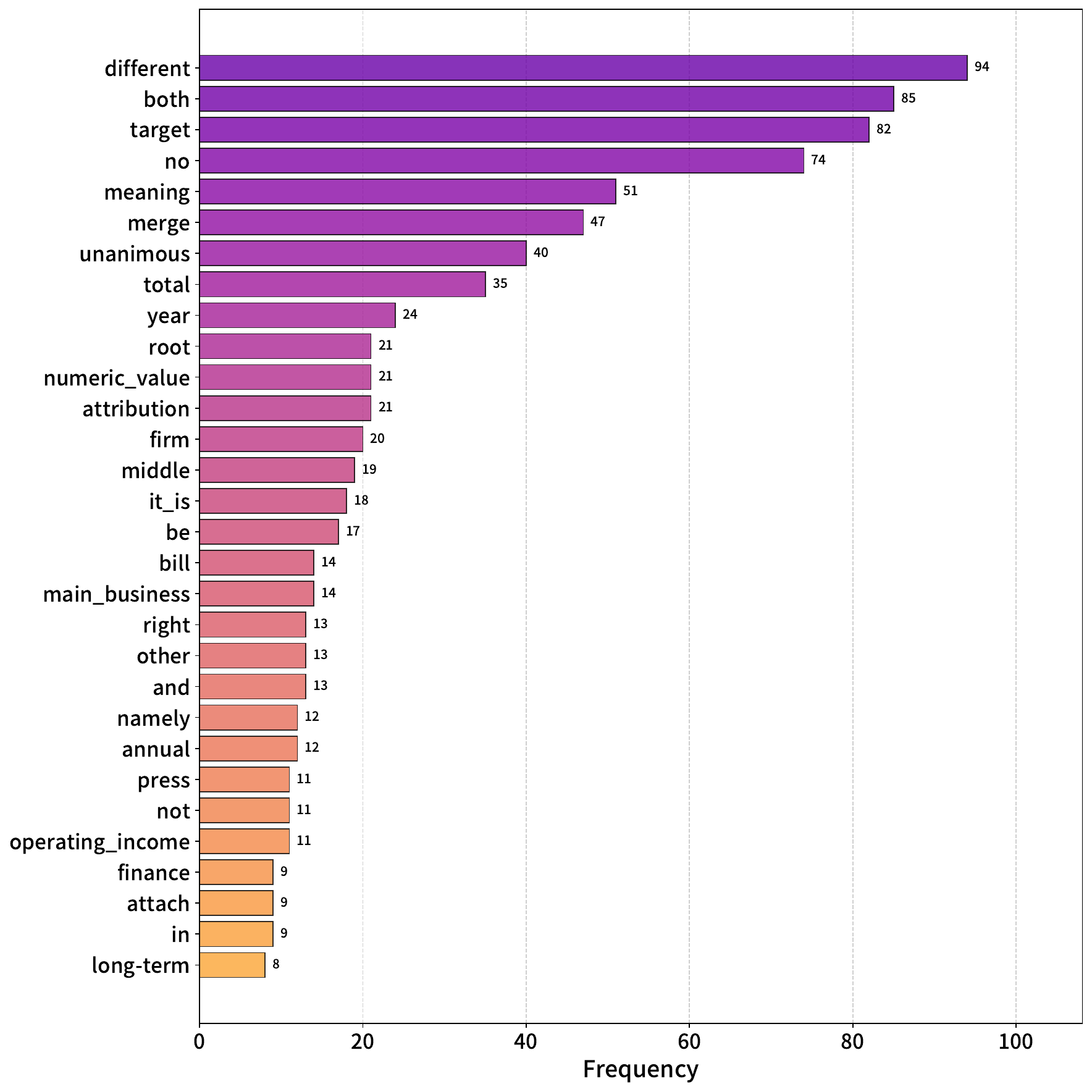}
        \caption{Top forking token frequencies for \textbf{SFT+RLVR}.}
        \label{fig:sft_rft}
    \end{minipage}
    \hfill
    \begin{minipage}{0.32\textwidth}
        \centering
        \includegraphics[width=0.9\linewidth]{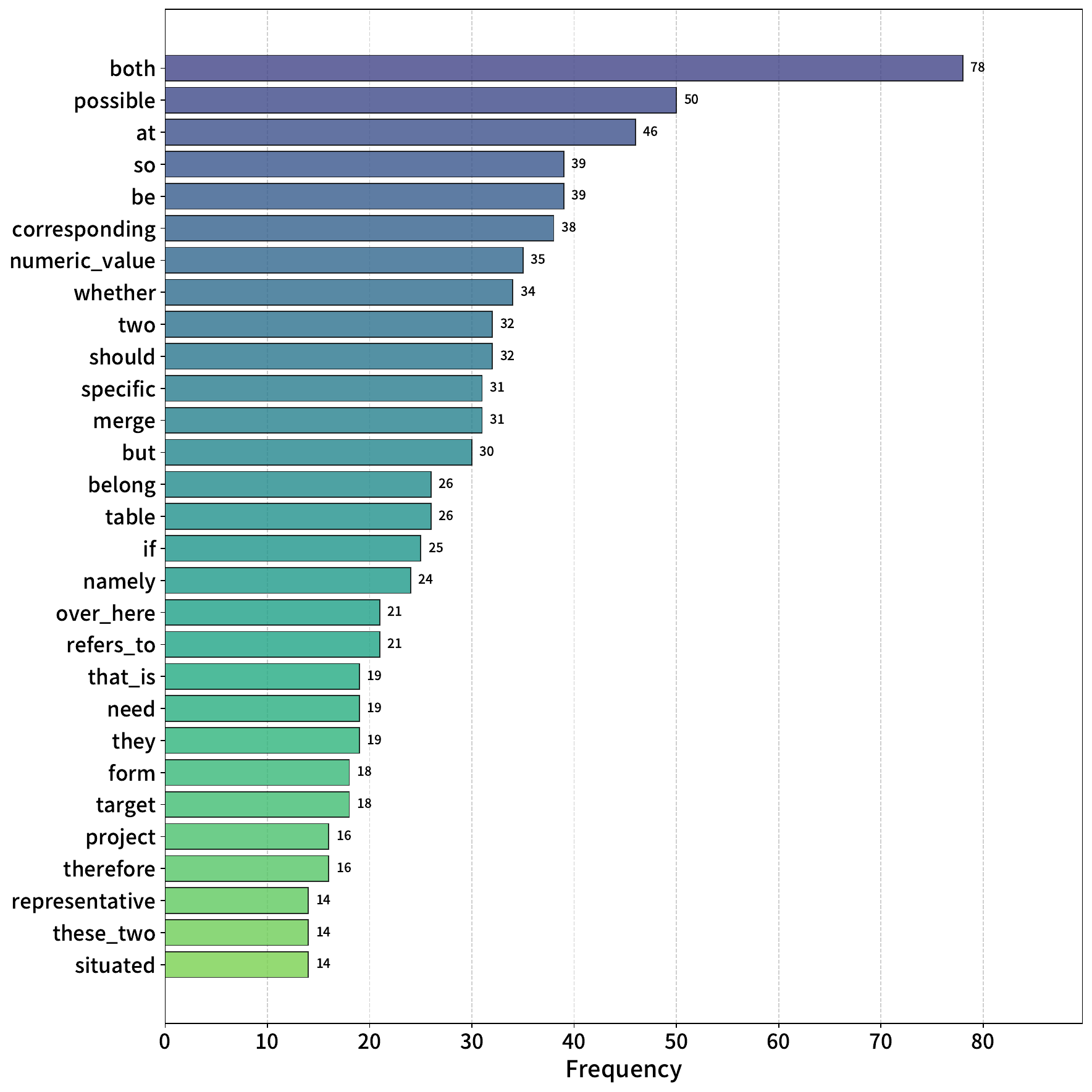}
        \caption{Top forking token frequencies for \textbf{UFT}.}
        \label{fig:curriculum_hint}
    \end{minipage}
    \label{fig:curr_hint}
    \hfill
    \begin{minipage}{0.32\textwidth}
        \centering
        \includegraphics[width=0.9\linewidth]{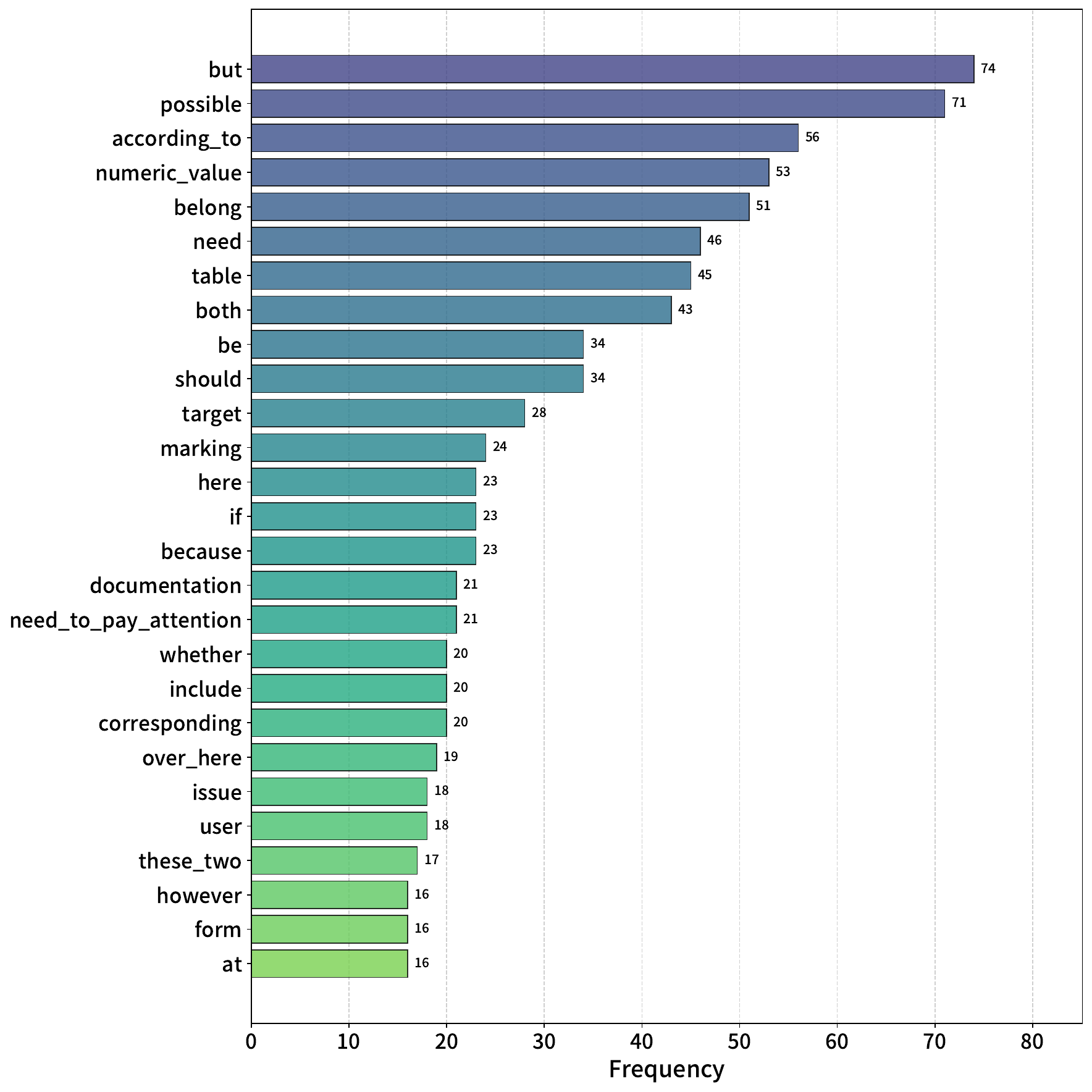}
        \caption{Top forking token frequencies for \textbf{pure-RLVR}.}
        \label{fig:warmup_hint}
    \end{minipage}
\end{figure*}

\subsubsection{Rollout-based Forking Token Detection}\label{sec:rftd}

Traditional entropy-based methods~\cite{wang2025beyond,cheng2025reasoning} for identifying forking tokens typically consider token positions with high entropy as forking tokens. However, these approaches suffer from a fundamental limitation: \textit{they misclassify positions where candidate tokens are semantically similar but probabilistically diverse as genuine forking tokens}. For instance, a position may contain synonymous candidate tokens such as \texttt{company} and \texttt{company's}, which carry nearly identical semantic meaning and exhibit high entropy, yet do not represent a critical decision point in the reasoning process. We present a detailed case study in Appendix B. This leads to the erroneous identification of positions that do not substantially influence the reasoning trajectory, thereby reducing the accuracy of forking token detection. To address this limitation, we propose \textbf{Rollout-based Forking Token Detection (RFTD)}, a novel approach that transcends probability-based heuristics by empirically evaluating the actual downstream impact of token substitutions through controlled generation rollouts.

The RFTD method operates by first identifying candidate positions based on entropy, then empirically testing their importance through token substitutions and continuation generation. Let $R = \{t_1, t_2, \ldots, t_L\}$ denote a response sequence of length $L$. We define the entropy at position $i$ as $H(t_i) = -\sum_{v \in V} P(v|t_1, \ldots, t_{i-1}) \log P(v|t_1, \ldots, t_{i-1})$, where $V$ is the vocabulary. For each candidate position, we substitute it with alternative tokens and measure the proportion of rollouts that produce significantly different outcomes. A position is classified as a forking token if this proportion exceeds threshold $\alpha$.

\begin{algorithm}[t]
\caption{Rollout-based Forking Token Detection (RFTD)}
\label{alg:rftd}
\begin{algorithmic}[1]
\REQUIRE Response $R = \{t_1, \ldots, t_L\}$, probability distributions $\{P_i\}_{i=1}^L$, threshold $\alpha$, hyperparameters $k, m, n$
\ENSURE Forking token set $F$
\STATE $F \leftarrow \emptyset$ \COMMENT{Initialize empty forking token set}
\STATE $T \leftarrow \text{TopK}(\{H(P_i) \mid i \in [1, L]\}, k)$ \COMMENT{Select top-$k$ positions with highest entropy}
\FOR{$t_{c} \in T$}
    \STATE $S \leftarrow \text{TopK}(\{P(v|t_1, \ldots, t_{c-1}) \mid v \in V \setminus \{t_{c}\}\}, m)$ \COMMENT{Select top-$m$ alternative tokens}
    \FOR{$s_j \in S$}
        \STATE $R' \leftarrow R[:c-1] \oplus s_j$ \COMMENT{Replace original token with substitute}
        \STATE $\{C_1, C_2, \ldots, C_n\} \leftarrow \text{Generate}(R', n)$
        \STATE $\rho_j \leftarrow \frac{1}{n}\sum_{l=1}^{n} \text{Divergent}(C_l, R)$ 
    \ENDFOR
    \IF{$\max_{j} \rho_j > \alpha$}
        \STATE $F \leftarrow F \cup \{t_{c}\}$ \COMMENT{Add to forking token set}
    \ENDIF
\ENDFOR
\RETURN $F$
\end{algorithmic}
\end{algorithm}
The algorithm is presented in Algorithm~\ref{alg:rftd}, where $\text{Generate}(R', n)$ prompts the LLM to produce $n$ continuation sequences from prefix $R'$, and $\text{Divergent}(C, R)$ returns 1 if the final answer of continuation $C$ differs from response $R$, 0 otherwise. This approach effectively distinguishes between tokens that merely exhibit high uncertainty in the probability distribution and those that genuinely influence the subsequent reasoning trajectory, thereby providing more accurate identification of critical decision points in responses.

\subsubsection{Analysis of Forking Tokens}\label{sec:analysis_fork}

In this section, we employ our RFTD method to analyze and compare the reasoning behaviors of models trained with different learning strategies. Specifically, we compare \textbf{SFT+RLVR} with other reasoning-enhancing methods such as \textbf{pure-RLVR} and \textbf{UFT}, using the full test set. Detailed experimental settings are presented in Appendix~B. We identify and visualize the frequency of forking tokens for these models in Figure~\ref{fig:sft_rft}–\ref{fig:warmup_hint}, revealing clear differences in their reasoning behaviors.

\textbf{SFT+RLVR produces task-relevant forking tokens.}  
The forking tokens of SFT+RLVR show strong alignment with the reasoning pattern of the NSM task. Tokens such as \textit{different}, \textit{unanimous}, \textit{not}, and \textit{no} directly correspond to semantic equivalence judgment, while others like \textit{annual}, \textit{firm}, \textit{main\_business}, and \textit{operating\_income} reflect reasoning about the contextual meaning of numerical values. 

\textbf{Other methods exhibit meta-reasoning or exploratory behaviors.}  
In contrast, models trained with alternative strategies (e.g., UFT, pure-RLVR) display forking tokens dominated by logical connectors and meta-reasoning indicators such as \textit{but}, \textit{according\_to}, \textit{if}, \textit{because}, and \textit{possible}. These tokens suggest that such models tend to rely on generic reasoning behaviors like hypothesis generation (\textit{if}, \textit{possible}), causal explanation (\textit{because}), and discourse linking (\textit{but}, \textit{however}) rather than task-specific reasoning.

Overall, these observations demonstrate that \textbf{SFT+RLVR more effectively internalizes the reasoning pattern required by the NSM task}.
By explicitly learning from rationale supervision, SFT+RLVR encourages the model to focus on these essential steps, leading to more accurate and interpretable reasoning behavior.

%% file: application.tex
\section{Application: Pattern-Aware LLMs as Rationale Annotators}

Building on the insights from the above controlled experiments and analysis, we introduce \textbf{P}attern-\textbf{A}ware LLMs as \textbf{R}ationale Ann\textbf{O}tators (\textbf{PARO}), a framework designed to reduce the rationale annotation cost in SFT+RLVR. 
Motivated by our finding that reasoning patterns are more critical than the quantity or quality of rationales, PARO aims to lessen the dependence on large-scale human-annotated rationale datasets by leveraging LLMs to automatically generate high-quality rationales in place of costly human annotation.

Specifically, we prompt state-of-the-art LLMs with reasoning pattern priors to synthesize rationales for given (question, answer) pairs. The reasoning pattern prior is encoded within the prompt instructions. For each task, we provide step-wise reasoning guidance along with two manually-annotated exemplar rationales to guide the model's reasoning process. Importantly, we do not provide the final answer as part of the prompt to prevent the model from generating shortcut rationales that bypass proper reasoning.

\subsection{Experimental Setup}\label{sec:paro_experiments}

We apply PARO to two representative reasoning tasks: Numerical Semantic Matching (NSM) and Transaction Purpose Classification (TPC), as detailed in Section~\ref{sec:nsm} and Section~\ref{sec:tpc}. 
We evaluate four rationale annotation strategies: 
(1) \textbf{SFT(1k, Human)+RLVR}, where SFT is trained on 1k human-annotated rationales; 
(2) \textbf{SFT(10k, Human)+RLVR}, which uses the full human-annotated dataset (10k samples for NSM only); 
(3) \textbf{SFT(1k, Distill)+RLVR}, where rationales are distilled directly from the internal reasoning traces of a large reasoning model; and 
(4) \textbf{SFT(1k, PARO)+RLVR}, where rationales are synthesized by PARO based on reasoning pattern priors.

To generate rationales, we employ Qwen3-235B-A22B-Thinking~\cite{qwen3technicalreport} (Qwen3-235B), a state-of-the-art open-source reasoning LLM, as the rationale annotator. We present the prompt template for these two tasks in Appendix E (Figure~\ref{fig:prompt-design} and Figure~\ref{fig:prompt-design-tpc}). 

For the NSM task, the dataset is the same as described in Section~\ref{sec:nsm_data}. For the TPC task, we collect 1k $(q, r, a)$ samples with manually annotated rationales for SFT training and 40k $(q, a)$ samples for RLVR training. The training recipes for both tasks follow the same configuration described in Section~\ref{sec:implement_detail}.

\subsection{Results and Analysis}

\begin{table}[!tbp]
\centering
\begin{tabular}{lcccc}
\toprule
\textbf{Strategy} & \textbf{Acc.} & \textbf{P.} & \textbf{R.} & \textbf{F1} \\
\midrule
\multicolumn{5}{l}{\textit{Numerical Semantic Matching (NSM)}} \\
SFT(1k, Human)+RLVR & 91.8 & 85.2 & 79.5 & 82.3 \\
SFT(10k, Human)+RLVR & \textbf{92.3} & \textbf{87.1} & 79.6 & 83.2 \\
SFT(1k, Distill)+RLVR & 90.4 & 83.1 & 76.0 & 79.4 \\
SFT(1k, PARO)+RLVR & 92.2 & 84.4 & \textbf{82.9} & \textbf{83.6} \\
\midrule
\multicolumn{5}{l}{\textit{Transaction Purpose Classification (TPC)}} \\
SFT(1k, Human)+RLVR & 87.9 & 87.6 & 87.9 & 87.2 \\
SFT(1k, Distill)+RLVR & 85.7 & 86.9 & 85.7 & 85.6 \\
SFT(1k, PARO)+RLVR & \textbf{88.2} & \textbf{89.0} & \textbf{88.2} & \textbf{87.9} \\
\bottomrule
\end{tabular}
\caption{Performance comparison of rationale annotation strategies on NSM and TPC tasks.}
\label{tab:paro_results}
\end{table}

As shown in Table~\ref{tab:paro_results}, PARO achieves strong and consistent performance across both tasks. On the NSM dataset, \textbf{SFT(1k, PARO)+RLVR} attains an accuracy of 92.2 and F1 of 83.6, closely matching the fully human-annotated \textbf{SFT(10k, Human)+RLVR} (92.3 accuracy, 83.2 F1), while reducing annotation cost by an order of magnitude without human supervision. Similarly, on the TPC task, PARO not only surpasses the distillation-based baseline by 2.5 accuracy points and 2.3 F1 points, but also slightly outperforms the human-annotated counterpart. 

These results confirm that incorporating reasoning pattern priors allows LLMs to produce concise, structured, and functionally equivalent rationales—mitigating the verbosity and noise often found in raw reasoning traces of LLMs. Overall, PARO effectively bridges the gap between human and LLM-generated rationales, offering a scalable and cost-efficient pathway for reasoning supervision in LLMs.

%% file: related_work.tex
\section{Related Work}

\subsection{Reinforcement Learning with Verifiable Rewards}
Reinforcement Learning with Verifiable Rewards (RLVR)~\cite{Lambert2024Tulu3RLVR} has proven highly effective for enhancing the reasoning abilities of LLMs across diverse domains. A key advantage of RLVR lies in its ability to optimize models using rule-based rewards derived solely from verifiable (question, answer) pairs, eliminating the need for costly human-annotated reasoning trajectories and thereby enabling scalable training on large datasets~\cite{guo2025deepseek}. In mathematical reasoning~\cite{shao2024deepseekmath}, such rewards are typically defined through symbolic or numerical verification rules, whereas in code generation~\cite{guo2025deepseek,jiang2024survey}, executable program outputs provide objective feedback signals.
Prior to the RLVR stage, models are typically warmed up through Supervised Fine-Tuning (SFT) on human-annotated reasoning trajectories (rationales) to establish initial reasoning behaviors~\cite{guo2025deepseek}. 
Recent research has explored various strategies for integrating human reasoning trajectories into the RLVR framework. Some efforts aim to enhance interpretability by generating human-readable reasoning paths~\cite{guo2025deepseek}, whereas others attempt to unify supervised and reinforcement learning under joint optimization frameworks~\cite{xi2024training,liu2025uft,fu2025srft}. However, the substantial cost of annotating large-scale, high-quality rationales remains largely overlooked. In contrast, we identify a broad class of problems, \textit{patterned reasoning tasks}, for which rationale annotation can be reliably automated by LLMs without compromising final performance.


\subsection{Reasoning Pattern Analysis}
Understanding how models learn and apply reasoning patterns has recently attracted significant research attention. 
Recent work on forking token analysis~\cite{wang2025beyond,cheng2025reasoning} has provided valuable insights into the mechanisms underlying model reasoning behavior. 
However, these studies typically employ entropy-based detection methods, treating high-entropy tokens as forking tokens. 
Such approaches may misclassify positions where candidate tokens are semantically similar but probabilistically diverse as genuine forking points, potentially leading to inaccurate analyses of reasoning patterns. 
Recent research~\cite{matsutani2025rl} has also explored how SFT and RL jointly shape reasoning behaviors by analyzing reasoning trajectories and constructing reasoning graphs, revealing that SFT expands correct reasoning paths while RL compresses incorrect ones, concentrating reasoning functionality into fewer steps. 
While their study focuses on how training dynamics restructure reasoning processes, our work instead investigates, under the SFT+RL paradigm, how the reasoning capabilities established during the SFT stage can be efficiently supervised, emphasizing the central role of reasoning patterns.

\subsection{Numerical Semantic Matching} 
Numerical semantic matching (NSM) has emerged as a critical task with significant overlap with document understanding~\cite{ma2024mmlongbench} and information extraction~\cite{pang2023guideline}. Early approaches~\cite{li2020cracking} primarily focused on training models to perform binary classification directly, without explicitly modeling the underlying reasoning process. However, these approaches overlook the sophisticated reasoning capabilities required to understand and compare the multi-faceted semantics of numerical mentions. While recent work~\cite{pang2025document} has begun exploring more sophisticated LLM-based approaches for enhanced numerical understanding, the reasoning-intensive nature of NSM remains largely underexplored. This work uses NSM as a representative patterned reasoning task and provide the first exploration of the reasoning nature inherent in the NSM task. 

%% file: conclusion.tex
\section{Conclusion and Future Work}

This paper revisits the role of rationales in the standard SFT+RLVR training pipeline for large language models. 
Through a systematic analysis of patterned reasoning tasks such as numerical semantic matching, we demonstrate that reasoning patterns—rather than the quantity or quality of rationales—serve as the primary determinant of reasoning performance. 
Building on these insights, we introduce \textbf{PARO} (\textbf{P}attern-\textbf{A}ware LLMs as \textbf{R}ationale Ann\textbf{O}tators), a cost-efficient annotation framework that leverages pattern-aware LLMs to automatically generate high-quality rationales. 
PARO achieves comparable or even superior performance to models trained on large human-annotated datasets, while eliminating the need for human rationale annotation, thereby offering a practical and scalable approach to reasoning supervision in LLMs.

Looking ahead, several promising directions merit further exploration:  
(1) Generalization to broader reasoning domains. Extending PARO to tasks involving logical, temporal, or spatial reasoning~\cite{wan2024logicasker,liu2025logic,cheng2024spatialrgpt,ICLR2025_b633e705,xiong2024large,wang2024tram} could reveal the broader applicability of the reasoning-pattern perspective.  
(2) Automated discovery of reasoning patterns. Developing methods that automatically extract or infer reasoning structures from unlabeled data would further minimize human involvement and enable large-scale, self-improving reasoning supervision.  
(3) Adaptive reasoning supervision. Future work could investigate hybrid strategies that dynamically balance pattern enforcement with exploratory reasoning, bridging the gap between patterned reasoning and adaptive domains such as mathematics or code generation.

Overall, our findings highlight a paradigm shift: for patterned reasoning tasks, effective reasoning supervision lies not in more or better rationales, but in teaching models \emph{how to reason}—through explicit reasoning patterns.

%% file: appendix.tex
\section*{Appendixes}

\subsection*{Appendix A: NSM Dataset Details}\label{app_dataset}

\begin{figure}[!htb]
    \centering
    \begin{minipage}{0.48\linewidth}
        \centering
        \includegraphics[width=\linewidth]{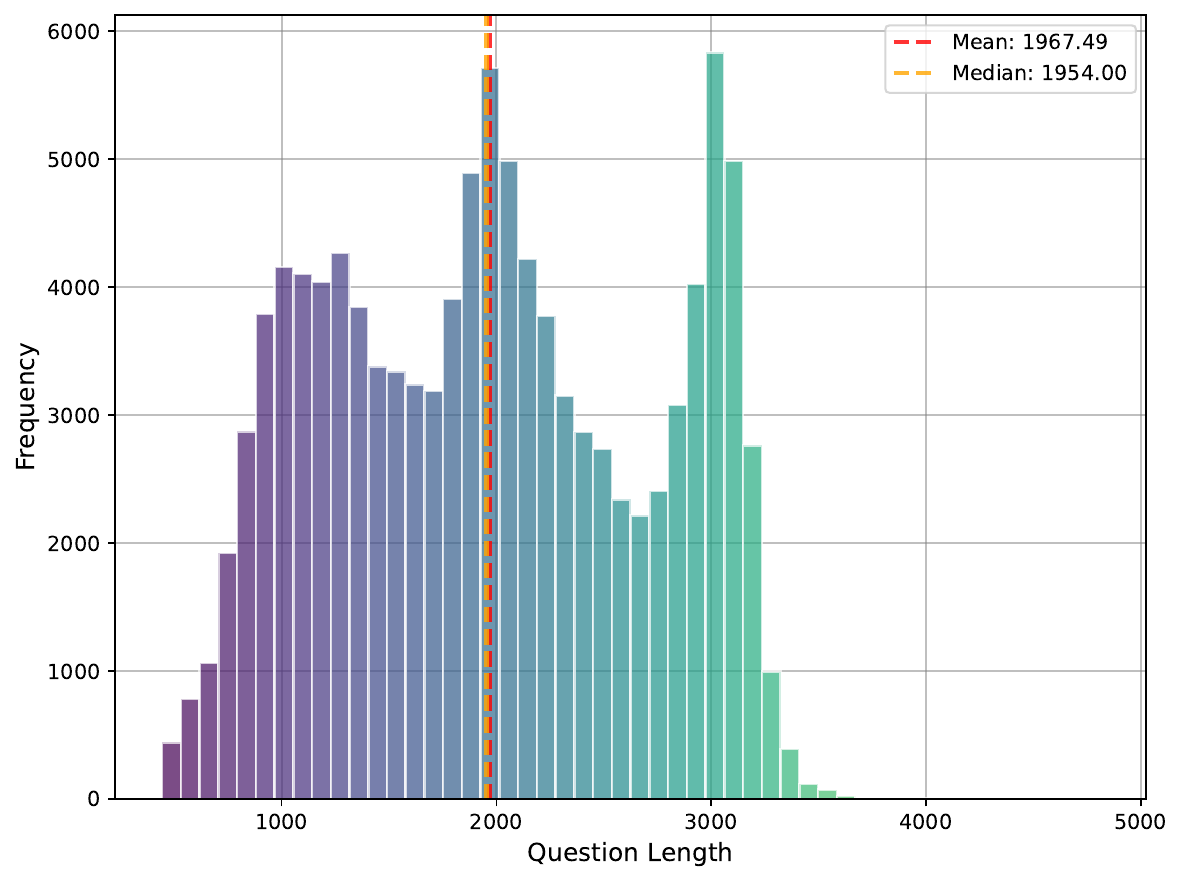}
        \caption{Frequency distribution of question lengths in the dataset.}
        \label{fig:ques_len}
    \end{minipage}
    \hfill
    \begin{minipage}{0.48\linewidth}
        \centering
        \includegraphics[width=\linewidth]{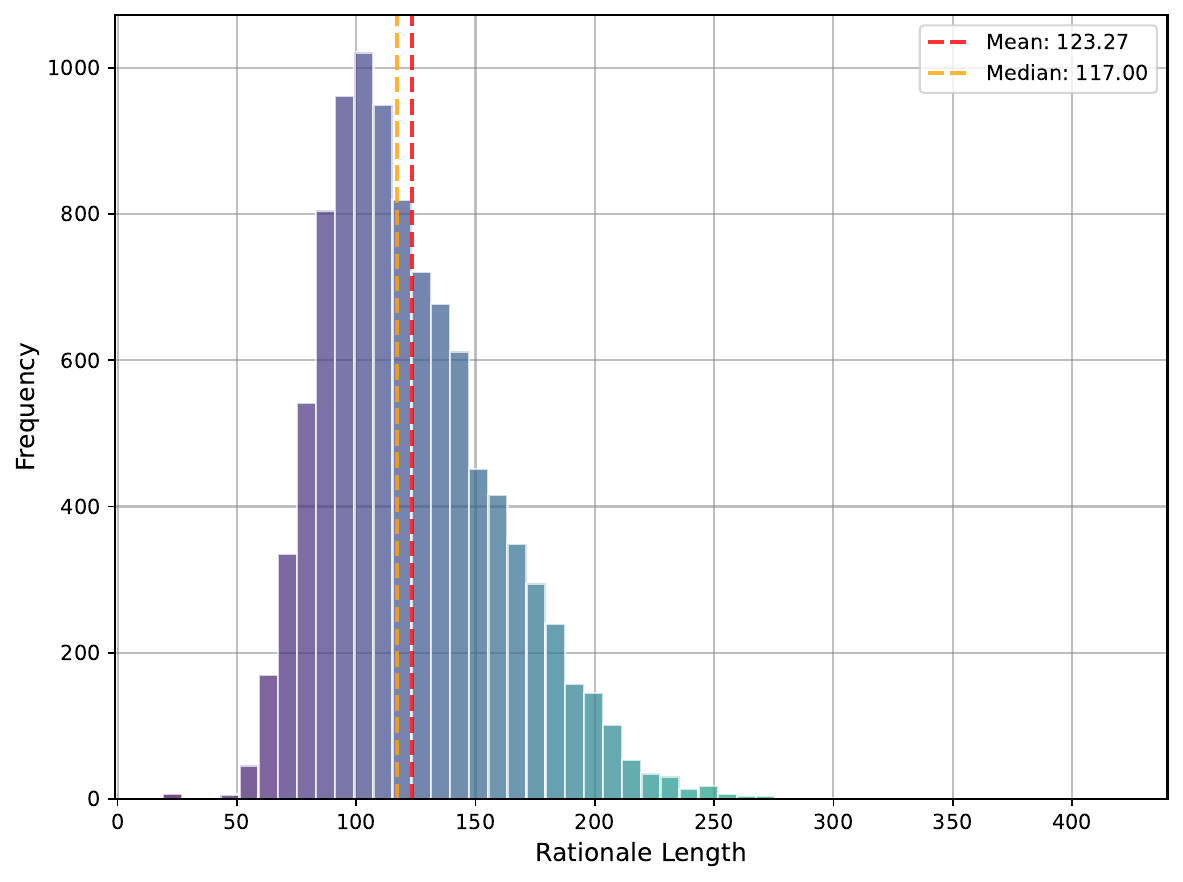}
        \caption{Frequency distribution of rationale lengths in the dataset.}
        \label{fig:rat_len}
    \end{minipage}
\end{figure}

\begin{table}[ht]
\centering
\normalsize
\begin{tabular}{lccc}
\toprule
\textbf{Context Type} & \textbf{Count} & \textbf{Ratio} \\
\midrule
Table-Table & 98{,}155 & 89.2\% \\
Table-Paragraph & 8{,}811 & 8.0\% \\
Paragraph-Paragraph & 3{,}034 & 2.8\% \\
\bottomrule
\end{tabular}
\caption{Statistics of context types for pairs of numerical mentions in each sample.}
\label{tab:context_distribution}
\end{table}

We present the frequency distributions of question lengths and rationale lengths in Figure~\ref{fig:ques_len} and Figure~\ref{fig:rat_len}, respectively. Length is measured by the number of tokens obtained after tokenizing each string with the Qwen3 tokenizer~\cite{yang2025qwen3}. Table~\ref{tab:context_distribution} summarizes the statistics of context types. The majority of samples fall under the Table-Table category, which is likely due to most numerical mentions appearing in tables rather than paragraphs.

\subsection*{Appendix B: Details of Forking Token Analysis}\label{appendix:exp_settings}
\textbf{Experimental details}. We use consistent experimental settings across pure-RLVR, UFT, and SFT+RLVR. To generate the original responses, we first employ the model to predict on the full test set using greedy decoding (temperature=0). We then apply the RFTD algorithm~\ref{alg:rftd} for forking token detection. The hyperparameters are set as follows: k=5, m=3, n=10, $\alpha=0.5$. For continuous generation, we use the corresponding LLM with a relatively high temperature of 0.7. Note that we report the chosen token at the forking position in the original response. We use vLLM~\cite{kwon2023efficient} for efficient inference.

\begin{figure}[!tb]
    \centering
    \includegraphics[width=0.9\linewidth]{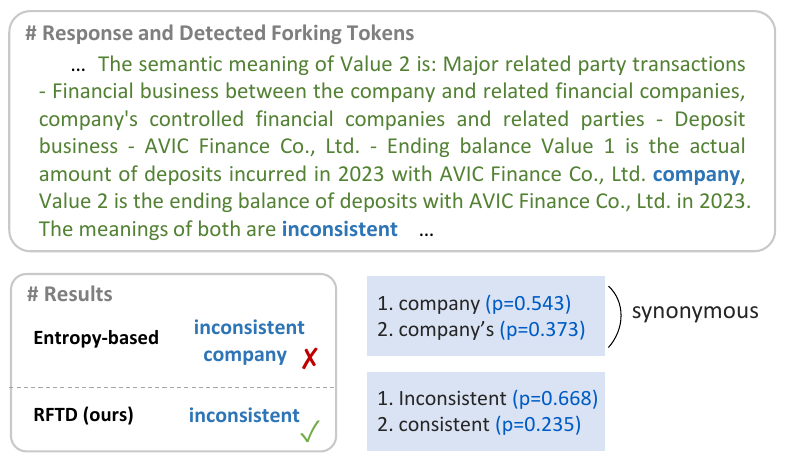}
    \caption{Case study: Traditional entropy-based methods incorrectly identify tokens with synonymous candidates as forking tokens. We display the top-2 candidate tokens for each position along with their corresponding probabilities.}
    \label{fig:case_study_synon}
\end{figure}

\noindent \textbf{Case study}. As presented in Figure~\ref{fig:case_study_synon}, we demonstrate the limitations of previous entropy-based methods~\cite{wang2025beyond,cheng2025reasoning} in accurately detecting forking tokens. These approaches typically identify token positions with high entropy in the output distribution as forking tokens. However, they frequently misclassify positions where candidate tokens are synonymous or semantically equivalent, such as when distinguishing between \texttt{company} and \texttt{company's}. In contrast, our rollout-based forking token detection method empirically evaluates the actual downstream impact of token substitutions through multiple rollouts, making it more robust to synonymous equivalence.

\subsection*{Appendix C: Prompts}\label{app:prompt}

We provide the detailed instructions for generating incorrect rationales in Figure~\ref{fig:task_instructions}.

\begin{figure}[!t]
\centering
\fbox{\begin{minipage}{0.45\textwidth}
\textbf{\# Task Instructions}

\textcolor{blue}{\{task instruction\}}

 Please note that this sample provides manually annotated hints before the Output. You may refer to the Hint content, but be aware that the Hint may not be complete.

\textbf{\# Input}

\textcolor{blue}{\{input\}}

\textbf{\# Hint}

\textcolor{blue}{\{hint\}}

\textbf{\# Output}

\end{minipage}}
\caption{Prompt template for the numerical semantic matching task with hint guidance.}
\label{fig:prompt_nsm_hint}
\end{figure}

\begin{figure}[!t]
\centering
\fbox{\begin{minipage}{0.45\textwidth}
\textbf{\# Task Instructions}

Given a Question and an Answer, please output a \texttt{Modified Answer} that transforms the Answer into an incorrect response.

Requirements:
\begin{itemize}
    \item Ensure that the final answer in \texttt{Modified Answer} is \textbf{different} from the original Answer.
    \item If the Answer's final conclusion is ``yes'', then the \texttt{Modified Answer}'s final conclusion should be ``no'', and vice versa.
    \item Output the \texttt{Modified Answer} directly without additional explanations.
\end{itemize}

\textbf{\# Question}

\textcolor{blue}{\{question\}}

\textbf{\# Answer}

\textcolor{blue}{\{answer\}}

\textbf{\# Modified Answer}

\end{minipage}}
\caption{Prompt template for generating incorrect rationales}
\label{fig:task_instructions}
\end{figure}

\subsection*{Appendix D: Details of the TPC Task}

\paragraph{Task Definition}  
The \textit{Transaction Purpose Classification} (TPC) task aims to classify a single bank transaction into one of 62 predefined purpose categories, including 42 corporate-related and 20 personal-related categories. Each transaction record provides multiple fields: account holder, transaction direction, transaction memo, counterparty, and contextual information. The objective is to predict the correct purpose category based on these inputs.

Each input is a structured record containing the following fields:
\begin{itemize}
    \item \textbf{Account holder type:} Enterprise or Individual
    \item \textbf{Transaction direction:} Credit or Debit
    \item \textbf{Transaction memo:} Free-text description of the transaction
    \item \textbf{Counterparty:} The entity on the other side of the transaction
    \item \textbf{Contextual cues:} Additional metadata such as time, channel, or amount
\end{itemize}

The expected output is:
\begin{itemize}
    \item A \textbf{category label} chosen from the taxonomy (62 classes in total).
\end{itemize}

\paragraph{Illustrative Example.}
\begin{quote}
\textbf{Input:}\\
Account holder: Xima Intelligent Technology Co., Ltd. \\
Account number: 88010122000085759 \\
Transaction date: 2019-03-11 \\
Credit: 10,000,000.0 \\
Debit: 0.0 \\
Balance: 10,018,196.76 \\
Transaction memo: ``Structured deposit principal'' \\
Counterparty: Xima Intelligent Technology Co., Ltd. \\
Counterparty account: 88010122000173077

\textbf{Output:}\\
Label: \texttt{Non-operating Income--Other Income} 
\end{quote}

\paragraph{Example Subset of Labels}  
Since the full taxonomy contains 62 categories, we list a representative subset here:
\begin{itemize}
    \item \texttt{Corporate--Tax Payment}
    \item \texttt{Corporate--Salary Distribution}
    \item \texttt{Corporate--Loan Repayment}
    \item \texttt{Corporate--Supplier Payment}
    \item \texttt{Personal--Credit Card Repayment}
    \item \texttt{Personal--Utility Bill Payment}
    \item \texttt{Personal--E-commerce Purchase}
    \item \texttt{Personal--Peer-to-Peer Transfer}
\end{itemize}

\subsection*{Appendix E: Prompt Templates of PARO}
We show the prompt template of PARO for the NSM and TPC task in Figure~\ref{fig:prompt-design} and Figure~\ref{fig:prompt-design-tpc}, respectively.

\begin{figure}[!tbp]
\centering
\fbox{\begin{minipage}{0.45\textwidth}
\textbf{Original Prompt:}
... If they are semantically equivalent, please output "yes", otherwise please output "no".

\textbf{Prompt with Reasoning Pattern Prior:}
... If they are semantically equivalent, please output "yes", otherwise please output "no". Please first provide your reasoning process in \texttt{<rationale>} and \texttt{</rationale>} tags, including: (1) Analyze the semantics of Value 1 and Value 2; (2) Compare the similarities and differences between their semantics in terms of time, subject, scope, entity, etc. If there is a difference in any aspect, then the output should be "no", otherwise output "yes". Please follow the format below for output and do not output any other content: \textcolor{blue}{\textbf{\{two-shot manually-annotated rationales\}}}
\end{minipage}}
\caption{Prompt design with reasoning pattern prior for NSM.}
\label{fig:prompt-design}
\end{figure}

\begin{figure}[!tbp]
\centering
\fbox{\begin{minipage}{0.45\textwidth}
\textbf{Original Prompt:}\\
... Please classify the purpose of the given bank transaction into one of the predefined categories.

\textbf{Prompt with Reasoning Pattern Prior:}\\
... Please classify the purpose of the given bank transaction into one of the predefined categories. 
Please first provide your reasoning process in \texttt{<rationale>} and \texttt{</rationale>} tags, following these structured steps:

\begin{enumerate}
    \item \textbf{Entity Identification:} Determine whether the account holder is an \textit{enterprise} (e.g., company, corporation) or an \textit{individual} (personal name).
    \item \textbf{Direction Determination:} Identify the transaction direction — whether it represents \textit{income (credit)} or \textit{expense (debit)}.
    \item \textbf{Information Matching:} Prioritize transaction keyword matching, then analyze the counterparty information:
    \begin{itemize}
        \item Financial institutions $\rightarrow$ investment / wealth management / loan categories
        \item Tax authorities $\rightarrow$ tax-related categories
        \item Judicial authorities $\rightarrow$ penalty / compensation categories
        \item Government departments $\rightarrow$ subsidy / tax-related categories
    \end{itemize}
    \item \textbf{Refined Classification:} Combine the subject type and transaction nature to select the most appropriate purpose category.
\end{enumerate}

Please follow the output format below and do not output any other content:\\
\textcolor{blue}{\textbf{\{two-shot manually-annotated rationales\}}}
\end{minipage}}
\caption{Prompt design with reasoning pattern prior for TPC.}
\label{fig:prompt-design-tpc}
\end{figure}

%% file: template.bbl
\begin{thebibliography}{10}

\bibitem{shao2024deepseekmath}
Shao Z, Wang P, Zhu Q, Xu~R, Song J, Bi~X, Zhang H, Zhang M, Li~Y, Wu~Y, others .
\newblock Deepseekmath: Pushing the limits of mathematical reasoning in open language models.
\newblock arXiv preprint arXiv:2402.03300, 2024

\bibitem{guo2025deepseek}
Guo D, Yang D, Zhang H, Song J, Zhang R, Xu~R, Zhu Q, Ma~S, Wang P, Bi~X, others .
\newblock Deepseek-r1: Incentivizing reasoning capability in llms via reinforcement learning.
\newblock arXiv preprint arXiv:2501.12948, 2025

\bibitem{chen2025bridging}
Chen H, Zheng K, Zhang Q, Cui G, Cui Y, Ye~H, Lin T~Y, Liu M~Y, Zhu J, Wang H.
\newblock Bridging supervised learning and reinforcement learning in math reasoning.
\newblock arXiv preprint arXiv:2505.18116, 2025

\bibitem{yu2025dapo}
Yu~Q, Zhang Z, Zhu R, Yuan Y, Zuo X, Yue Y, Dai W, Fan T, Liu G, Liu L, others .
\newblock Dapo: An open-source llm reinforcement learning system at scale.
\newblock arXiv preprint arXiv:2503.14476, 2025

\bibitem{jiang2024survey}
Jiang J, Wang F, Shen J, Kim S, Kim S.
\newblock A survey on large language models for code generation.
\newblock arXiv preprint arXiv:2406.00515, 2024

\bibitem{Lambert2024Tulu3RLVR}
Lambert N, Morrison J, Pyatkin V, Huang S, Ivison H, Brahman F, Miranda L~J~V, Liu A, Dziri N, Lyu S, Gu~Y, Malik S, Graf V, Hwang J~D, Yang J, Bras R~L, Tafjord O, Wilhelm C, Soldaini L, Smith N~A, Wang Y, Dasigi P, Hajishirzi H.
\newblock {T\"{U}LU} 3: Pushing frontiers in open language model post-training.
\newblock CoRR, 2024, abs/2411.15124

\bibitem{schulman2017proximal}
Schulman J, Wolski F, Dhariwal P, Radford A, Klimov O.
\newblock Proximal policy optimization algorithms.
\newblock arXiv preprint arXiv:1707.06347, 2017

\bibitem{position2025expensive}
Kandpal N, Raffel C.
\newblock Position: The most expensive part of an llm should be its training data.
\newblock arXiv preprint, 2025, arXiv:2504.12427

\bibitem{zhang2024sentiment}
Zhang W, others .
\newblock Sentiment analysis in the era of large language models.
\newblock In: Findings of the North American Chapter of the Association for Computational Linguistics (NAACL) -- Findings.
\newblock 2024.
\newblock Comprehensive evaluation of LLMs on sentiment tasks; discusses prompting and evaluation.

\bibitem{vykopal2024fact}
Vykopal I, others .
\newblock Generative large language models in automated fact-checking: A survey.
\newblock arXiv preprint, 2024.
\newblock Survey of LLM applications and limits in fact-checking, discussing model capabilities and reliance on human fact-checkers.

\bibitem{pang2023guideline}
Pang C, Cao Y, Ding Q, Luo P.
\newblock Guideline learning for in-context information extraction.
\newblock In: Proceedings of the 2023 Conference on Empirical Methods in Natural Language Processing.
\newblock 2023,  15372--15389

\bibitem{dagdelen2024structured}
Dagdelen J, others .
\newblock Structured information extraction from scientific text with pretrained large language models.
\newblock Nature Communications, 2024, 15: 45563.
\newblock Shows how LLMs can be fine-tuned for joint NER/RE and structured extraction in scientific domains.

\bibitem{zuo2025kg4diagnosis}
Zuo K, Jiang Y, Mo~F, Lio P.
\newblock Kg4diagnosis: A hierarchical multi-agent llm framework with knowledge graph enhancement for medical diagnosis.
\newblock In: AAAI Bridge Program on AI for Medicine and Healthcare.
\newblock 2025,  195--204

\bibitem{hillebrand2023improving}
Hillebrand L, Berger A, Deu{\ss}er T, Dilmaghani T, Khaled M, Kliem B, Loitz R, Pielka M, Leonhard D, Bauckhage C, others .
\newblock Improving zero-shot text matching for financial auditing with large language models.
\newblock In: Proceedings of the ACM Symposium on Document Engineering 2023.
\newblock 2023,  1--4

\bibitem{john2025human}
John L, Wittenborg T, Auer S, Karras O.
\newblock Human-in-the-loop workflow for neuro-symbolic scholarly knowledge organization.
\newblock arXiv preprint arXiv:2506.03221, 2025

\bibitem{hao2025realworldplanning}
Hao Y, Chen Y, Zhang Y, Fan C.
\newblock Large language models can solve real-world planning rigorously with formal verification tools.
\newblock In: Proceedings of the 2025 Conference of the Nations of the Americas Chapter of the Association for Computational Linguistics: Human Language Technologies (Long Papers).
\newblock 2025,  3434--3483

\bibitem{wang2025beyond}
Wang S, Yu~L, Gao C, Zheng C, Liu S, Lu~R, Dang K, Chen X, Yang J, Zhang Z, others .
\newblock Beyond the 80/20 rule: High-entropy minority tokens drive effective reinforcement learning for llm reasoning.
\newblock arXiv preprint arXiv:2506.01939, 2025

\bibitem{cheng2025reasoning}
Cheng D, Huang S, Zhu X, Dai B, Zhao W~X, Zhang Z, Wei F.
\newblock Reasoning with exploration: An entropy perspective.
\newblock arXiv preprint arXiv:2506.14758, 2025

\bibitem{liu2025uft}
Liu M, Farina G, Ozdaglar A.
\newblock Uft: Unifying supervised and reinforcement fine-tuning.
\newblock arXiv preprint arXiv:2505.16984, 2025

\bibitem{qwen3technicalreport}
Team Q.
\newblock Qwen3 technical report, 2025

\bibitem{zhang2024sentimentLLM}
Zhang W, Deng Y, Liu B, Pan S~J, Bing L.
\newblock Sentiment analysis in the era of large language models: A reality check.
\newblock In: Findings of the Association for Computational Linguistics: NAACL 2024.
\newblock 2024,  246–259

\bibitem{gretz2023zeroshot}
Gretz S, Halfon A, Shnayderman I, Toledo-Ronen O, Dankin L, Katsis Y, Arviv O, Katz Y, Slonim N, Ein-Dor L.
\newblock Zero-shot topical text classification with llms - an experimental study.
\newblock In: Findings of the Association for Computational Linguistics: EMNLP 2023.
\newblock 2023,  9647--9676

\bibitem{arora2024intentLLMs}
Arora G, Jain S, Merugu S.
\newblock Intent detection in the age of llms.
\newblock In: Proceedings of the 2024 Conference on Empirical Methods in Natural Language Processing (Industry Track).
\newblock 2024,  1559--1570

\bibitem{li2024self}
Li~M, Peng B, Galley M, Gao J, Zhang Z.
\newblock Self-checker: Plug-and-play modules for fact-checking with large language models.
\newblock In: Findings of the Association for Computational Linguistics: NAACL 2024.
\newblock 2024,  163--181

\bibitem{leippold2025automated}
Leippold M, Vaghefi S~A, Stammbach D, Muccione V, Bingler J, Ni~J, Senni C~C, Wekhof T, Schimanski T, Gostlow G, others .
\newblock Automated fact-checking of climate claims with large language models.
\newblock npj Climate Action, 2025, 4(1): 17

\bibitem{wan2023gpt}
Wan Z, Cheng F, Mao Z, Liu Q, Song H, Li~J, Kurohashi S.
\newblock Gpt-re: In-context learning for relation extraction using large language models.
\newblock arXiv preprint arXiv:2305.02105, 2023

\bibitem{pang2024uncovering}
Pang C, Cao Y, Yang C, Luo P.
\newblock Uncovering limitations of large language models in information seeking from tables.
\newblock In: Findings of the Association for Computational Linguistics ACL 2024.
\newblock 2024,  1388--1409

\bibitem{zhao2023investigating}
Zhao Y, Zhang H, Si~S, Nan L, Tang X, Cohan A.
\newblock Investigating table-to-text generation capabilities of large language models in real-world information seeking scenarios.
\newblock In: Proceedings of the 2023 Conference on Empirical Methods in Natural Language Processing: Industry Track.
\newblock 2023,  160--175

\bibitem{pan2023logic}
Pan L, Albalak A, Wang X, Wang W.
\newblock Logic-lm: Empowering large language models with symbolic solvers for faithful logical reasoning.
\newblock In: Findings of the Association for Computational Linguistics: EMNLP 2023.
\newblock 2023,  3806--3824

\bibitem{weng2025geosketch}
Weng S, Wang Z, Zhou Y, Lu~R, Liu T, Teng Z, Liu X, Liu H.
\newblock Geosketch: A neural-symbolic approach to geometric multimodal reasoning with auxiliary line construction and affine transformation, 2025

\bibitem{ma2024llm}
Ma~P, Wang T~H, Guo M, Sun Z, Tenenbaum J~B, Rus D, Gan C, Matusik W.
\newblock Llm and simulation as bilevel optimizers: a new paradigm to advance physical scientific discovery.
\newblock In: Proceedings of the 41st International Conference on Machine Learning.
\newblock 2024,  33940--33962

\bibitem{pang2025document}
Pang C, Cao Y, Zhou G, Li~H, Luo P.
\newblock Document-level tabular numerical cross-checking: A coarse-to-fine approach.
\newblock arXiv preprint arXiv:2506.13328, 2025

\bibitem{zheng2023survey}
Zheng H, Wang S, Huang L.
\newblock A survey of document-level information extraction.
\newblock arXiv preprint arXiv:2309.13249, 2023

\bibitem{ran2019numnet}
Ran Q, Lin Y, Li~P, Zhou J, Liu Z.
\newblock {NumNet}: Machine reading comprehension with numerical reasoning.
\newblock In: Proceedings of EMNLP.
\newblock 2019

\bibitem{akhtar2023exploring}
Akhtar M, others .
\newblock Exploring the numerical reasoning capabilities of language models: A comprehensive analysis on tabular data.
\newblock arXiv preprint arXiv:2311.02216, 2023

\bibitem{wang2024encore}
Wang D, others .
\newblock Enhancing numerical reasoning with the guidance of chain-of-thought (encore).
\newblock In: Proceedings of ACL (Long).
\newblock 2024

\bibitem{li2020cracking}
Li~H, Yang Q, Cao Y, Yao J, Luo P.
\newblock Cracking tabular presentation diversity for automatic cross-checking over numerical facts.
\newblock In: Proceedings of the 26th ACM SIGKDD International Conference on Knowledge Discovery \& Data Mining.
\newblock 2020,  2599--2607

\bibitem{yang2025qwen3}
Yang A, Li~A, Yang B, Zhang B, Hui B, Zheng B, Yu~B, Gao C, Huang C, Lv~C, others .
\newblock Qwen3 technical report.
\newblock arXiv preprint arXiv:2505.09388, 2025

\bibitem{galileo2025qwen_chinese_advantage}
Blog G.
\newblock Beyond gpt: How qwen is reshaping ai.
\newblock 2025.
\newblock “The inclusion of substantial Chinese-language content gives Qwen an advantage in understanding Chinese cultural contexts, idioms, and specialized terminology.”

\bibitem{wu2025thought}
Wu~J, Liao C, Feng M, Zhang S, Wen Z, Shao P, Xu~H, Tao J.
\newblock Thought-augmented policy optimization: Bridging external guidance and internal capabilities.
\newblock arXiv preprint arXiv:2505.15692, 2025

\bibitem{wolf2020transformers}
Wolf T, Debut L, Sanh V, Chaumond J, Delangue C, Moi A, Cistac P, Rault T, Louf R, Funtowicz M, others .
\newblock Transformers: State-of-the-art natural language processing.
\newblock In: Proceedings of the 2020 conference on empirical methods in natural language processing: system demonstrations.
\newblock 2020,  38--45

\bibitem{rajbhandari2020zero}
Rajbhandari S, Rasley J, Ruwase O, He~Y.
\newblock Zero: Memory optimizations toward training trillion parameter models.
\newblock In: SC20: International Conference for High Performance Computing, Networking, Storage and Analysis.
\newblock 2020,  1--16

\bibitem{sheng2024hybridflow}
Sheng G, Zhang C, Ye~Z, Wu~X, Zhang W, Zhang R, Peng Y, Lin H, Wu~C.
\newblock Hybridflow: A flexible and efficient rlhf framework.
\newblock arXiv preprint arXiv: 2409.19256, 2024

\bibitem{kwon2023efficient}
Kwon W, Li~Z, Zhuang S, Sheng Y, Zheng L, Yu~C~H, Gonzalez J, Zhang H, Stoica I.
\newblock Efficient memory management for large language model serving with pagedattention.
\newblock In: Proceedings of the 29th symposium on operating systems principles.
\newblock 2023,  611--626

\bibitem{jiang2025predicting}
Jiang C, Zhang M, Ye~J, Fan X, Cao Y, Sun J, Xi~Z, Dou S, Dong Y, Shen Y, Tong J, Wang Z, Liang T, Fei Z, Wan M, Ma~G, Zhang Q, Gui T, Huang X.
\newblock Predicting large language model capabilities on closed-book qa tasks using only information available prior to training.
\newblock arXiv preprint arXiv:2502.04066, 2025

\bibitem{min2022rethinking}
Min S, Lyu X, Holtzman A, Artetxe M, Lewis M, Hajishirzi H, Zettlemoyer L.
\newblock Rethinking the role of demonstrations: What makes in-context learning work?
\newblock In: EMNLP.
\newblock 2022

\bibitem{matsutani2025rl}
Matsutani K, Takashiro S, Minegishi G, Kojima T, Iwasawa Y, Matsuo Y.
\newblock Rl squeezes, sft expands: A comparative study of reasoning llms.
\newblock arXiv preprint arXiv:2509.21128, 2025

\bibitem{xi2024training}
Xi~Z, Chen W, Hong B, Jin S, Zheng R, He~W, Ding Y, Liu S, Guo X, Wang J, others .
\newblock Training large language models for reasoning through reverse curriculum reinforcement learning.
\newblock In: International Conference on Machine Learning.
\newblock 2024,  54030--54048

\bibitem{fu2025srft}
Fu~Y, Chen T, Chai J, Wang X, Tu~S, Yin G, Lin W, Zhang Q, Zhu Y, Zhao D.
\newblock Srft: A single-stage method with supervised and reinforcement fine-tuning for reasoning.
\newblock arXiv preprint arXiv:2506.19767, 2025

\bibitem{ma2024mmlongbench}
Ma~Y, Zang Y, Chen L, Chen M, Jiao Y, Li~X, Lu~X, Liu Z, Ma~Y, Dong X, others .
\newblock Mmlongbench-doc: Benchmarking long-context document understanding with visualizations.
\newblock Advances in Neural Information Processing Systems, 2024, 37: 95963--96010

\bibitem{wan2024logicasker}
Wan Y, Wang W, Yang Y, Yuan Y, Huang J~t, He~P, Jiao W, Lyu M.
\newblock Logicasker: Evaluating and improving the logical reasoning ability of large language models.
\newblock In: Proceedings of the 2024 Conference on Empirical Methods in Natural Language Processing.
\newblock 2024,  2124--2155

\bibitem{liu2025logic}
Liu T, Xu~W, Huang W, Zeng Y, Wang J, Wang X, Yang H, Li~J.
\newblock Logic-of-thought: Injecting logic into contexts for full reasoning in large language models.
\newblock In: Proceedings of the 2025 Conference of the Nations of the Americas Chapter of the Association for Computational Linguistics: Human Language Technologies (Volume 1: Long Papers).
\newblock 2025,  10168--10185

\bibitem{cheng2024spatialrgpt}
Cheng A~C, Yin H, Fu~Y, Guo Q, Yang R, Kautz J, Wang X, Liu S.
\newblock Spatialrgpt: grounded spatial reasoning in vision-language models.
\newblock In: Proceedings of the 38th International Conference on Neural Information Processing Systems.
\newblock 2024,  135062--135093

\bibitem{ICLR2025_b633e705}
Zhang Y, Xu~Z, Shen Y, Kordjamshidi P, Huang L.
\newblock Spartun3d: Situated spatial understanding of 3d world in large language model.
\newblock In: Yue Y, Garg A, Peng N, Sha F, Yu~R, eds, International Conference on Representation Learning.
\newblock 2025,  73388--73406

\bibitem{xiong2024large}
Xiong S, Payani A, Kompella R, Fekri F.
\newblock Large language models can learn temporal reasoning.
\newblock In: Proceedings of the 62nd Annual Meeting of the Association for Computational Linguistics (Volume 1: Long Papers).
\newblock 2024,  10452--10470

\bibitem{wang2024tram}
Wang Y, Zhao Y.
\newblock Tram: Benchmarking temporal reasoning for large language models.
\newblock In: Findings of the Association for Computational Linguistics ACL 2024.
\newblock 2024,  6389--6415

\end{thebibliography}
